\documentclass[twocolumn]{autart}    

\usepackage{graphicx}          

\usepackage{etex}        
\usepackage{amsfonts}
\usepackage{times}
\usepackage{latexsym,amsfonts,amsbsy,amssymb}
\usepackage{amsmath}
\usepackage{amssymb}
\usepackage{textcomp}
\usepackage{gensymb}

\usepackage{algorithm}
\usepackage{algpseudocode}
\usepackage{listings}
\newtheorem{remark}{Remark}
\newtheorem{corollary}{Corollary}[thm]
\newtheorem{problem}{Problem}
\theoremstyle{exampstyle} \newtheorem{example}{Example}
\algnewcommand\algorithmicinput{\textbf{INPUT:}}
\algnewcommand\INPUT{\item[\algorithmicinput]}

\usepackage{tabularx,booktabs}
\usepackage{epstopdf}
\usepackage{graphicx}
\usepackage{subfig}
\usepackage{caption}
\usepackage{color}
\usepackage{cite}
\usepackage{float}
\usepackage{tikz}
\usetikzlibrary{automata,positioning,shapes,arrows}

\begin{document}
	
\begin{frontmatter}
	
	\title{Combined Top-Down and Bottom-Up Approaches to Performance-guaranteed Integrated Task and Motion Planning of Cooperative Multi-agent Systems\thanksref{footnoteinfo}\thanksref{support}} 
	
	\thanks[footnoteinfo]{This paper was not presented at any IFAC 
		meeting. Corresponding author R.~R.~da~Silva. Tel. +1-574-631-3736. 
		Fax +1-574-631-4393.}
	
	\author[All]{Rafael~Rodrigues~da~Silva\thanksref{capes}}\ead{rrodri17@nd.edu},    
	\author[All]{Bo~Wu}\ead{bwu3@nd.edu},               
	\author[All]{Jin~Dai}\ead{jdai1@nd.edu},               
	\author[All]{Hai~Lin}\ead{hlin1@nd.edu}               
	
	\thanks[support]{This work is supported by NSF-CNS-1239222, NSF-EECS-1253488 and NSF-CNS-1446288}
	\thanks[capes]{The first author would like to appreciate the scholarship support by CAPES/BR, BEX 13242/13-0}
	
	\address[All]{Department of Electrical Engineering, University of Notre Dame, Notre Dame, IN, 46556 USA.}  


\begin{abstract}                
	We propose a hierarchical design framework to automatically synthesize coordination schemes and control policies for cooperative multi-agent systems to fulfill formal performance requirements, by associating a bottom-up reactive motion controller with a top-down mission plan. On one hand, starting from a global mission that is specified as a regular language over all the agents' mission capabilities, a mission planning layer sits on the top of the proposed framework, decomposing the global mission into local tasks that are in consistency with each agent's individual capabilities, and compositionally justifying whether the achievement of local tasks implies the satisfaction of the global mission via an assume-guarantee paradigm. On the other hand, bottom-up motion plans associated with each agent are synthesized corresponding to the obtained local missions by composing basic motion primitives, which are verified safe by differential dynamic logic (d$\mathcal{L}$), through a Satisfiability Modulo Theories (SMT) solver that searches feasible solutions in face of constraints imposed by local task requirements and the environment description. It is shown that the proposed framework can handle dynamical environments as the motion primitives possess reactive features, making the motion plans adaptive to local environmental changes. Furthermore, on-line mission reconfiguration can be triggered by the motion planning layer once no feasible solutions can be found through the SMT solver. The effectiveness of the overall design framework is validated by an automated warehouse case study.

	
	
\end{abstract}
	
\end{frontmatter}

%
%
%
%
%
%
%
%
%
%

%
%

\section{Introduction}

Cooperative multi-agent systems refer to as a class of multi-agent systems in which a number of homogeneous and/or heterogeneous agents collaborating in a distributed manner via wireless communication channels in order to accomplish desirable performance objectives cooperatively. Representing a typical class of cyber-physical systems (CPS), cooperative multi-agent systems has become a powerful analysis and design tool in the interdisciplinary study of control theory and computer science due to the great potential in both academia and industry, ranging from traffic management systems, power grids, robotic teams to smart manufacturing systems, see e.g. \cite{arkin1998behavior,chosetprinciples,fainekos2009temporal,lin2014mission,negenborn2008multi} and the references therein.

Mission and motion planning are two fundamental problems in the context of cooperative multi-agent systems and have received considerable attention in recent years. To pursue satisfaction of desired performance requirements, planning methods for cooperative multi-agent systems can generally be divided into two categories: {\it bottom-up} and {\it top-down} approaches. Bottom-up approaches design local control rules and inter-agent coordination mechanisms to fulfill each agent's individual tasks, while sophisticated collective behavior of cooperative multi-agent systems manages to ensure certain global properties. Such approaches have gained remarkable success in achieving various mission and motion planning purposes, including behavior-based coordination \cite{arkin1998behavior}, consensus-type motion planning \cite{olfati2004consensus} and local high-level tasks \cite{filippidis2012decentralized}. The bottom-up approach scales well but generally lacks formal performance guarantees, except for certain properties like consensus \cite{olfati2004consensus}, rendezvous \cite{dimarogonas2007rendezvous} or related formation control \cite{fax2004information}. In contrary, starting from a global mission, {\it top-down} design methods complements bottom-up ones by following a ``divide-and-conquer" paradigm, in which the global mission is decomposed into a series of local tasks for each agent based on their individual sensing and actuating capabilities, and accomplishment of the local missions ensures the satisfaction of the global specification via synchronized \cite{karaman2011linear} \cite{chen2012formal} or partially-synchronized \cite{kloetzer2010automatic}\cite{ding2011automatic} multi-agent coordination. Despite the guarantee of achieving complex high-level global mission and motion plans, top-down design methods lack flexibility and scalability in local control policy design due to their requirement for proper abstraction models. Additionally, the planning complexity quickly becomes prohibitively high as the number of partitioned
regions and agents increase, which further hampers the applicability of the abstraction based methods in many practical circumstances.

\begin{figure}[t]
	\centering
	\includegraphics[scale=0.7]{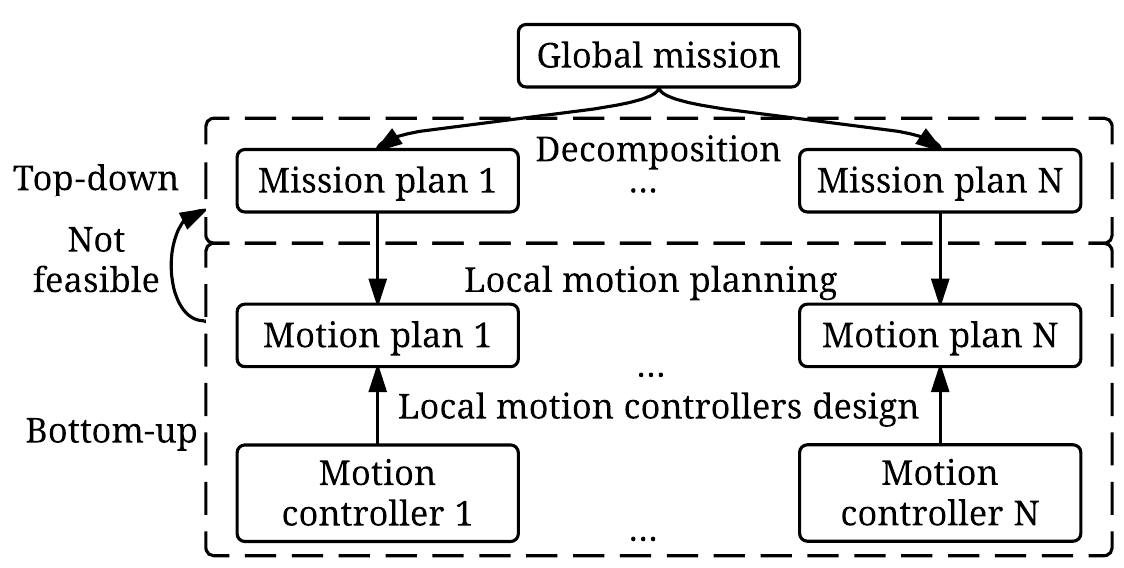}
	\caption{Overall framework}
	\label{fig:overallframework}
\end{figure}

We are therefore motivated to combine both top-down mission planning procedure with bottom-up motion planning techniques to develop a scalable, reactive and correct-by-design approach for cooperative multi-agent systems that accomplishes high-level global tasks in uncertain and dynamic environments. Our basic idea in this paper is illustrated by the design framework shown in Fig.~\ref{fig:overallframework}.

Given a global mission in the form of regular languages over the mission capabilities of the underlying cooperative multi-agent system, our proposed framework solves the mission planning problem by introducing a learning-based top-down mission decomposition framework \cite{dai2014automatic}, which decomposes the global mission into local tasks that are consistent with each agent's capabilities. Based on the given local tasks, we solve the corresponding integrated task and motion planning problem of the multi-agent systems by extending our previous results of bottom-up compositional design approach called CoSMoP (Composition of Safe Motion Primitives) \cite{CoSMoP} from single agent to multi-agent circumstances. First, CoSMoP designs a series of motion controllers (primitives) offline that are verified safe by differential dynamic logic (d$\mathcal{L}$) \cite{platzer2010logical} to form necessary building blocks of complex maneuvers for each agent. Next, with the learned local task specification and a scenario map, CoSMoP synthesizes the corresponding integrated task and motion plan via appropriate composition of simple motion primitives whose correctness is justified by using the Satisfiability Modular Theories (SMT) solver and by modular incremental verification procedures. The mission and motion planning problem can be solved successfully if the motion planning layer comes up with a set of feasible motion plans for a fair dynamic environment, i.e. the changes in the environment do not lead any agent to a deadlock. However, if it fails to obtain feasible motion plans or an agent gets in a deadlock, feedbacks can be provided to adjust each agent's mission plan by exploiting necessary coordinations. Our main contributions lie in
\begin{enumerate}
	\item We apply formal methods to solve both the mission and the motion planning problems of cooperative multi-agent systems, based on which provably correct mission plans are obtained and feasible motion plans are synthesized through correct-by-construction.
	\item Our proposed framework shows great improvement of the scalability issues. On one hand, in the top-down mission planning stage, we use assume-guarantee paradigm \cite{puasuareanu2008learning} to compositionally verify the correctness of all the mission plans, mitigating the ``state explosion" issues; on the other hand, we synthesize the corresponding motion controllers by using SMT solver and thus finite abstractions of the environment \cite{belta2007symbolic} is avoided.
	\item Although the given global mission is not necessarily reactive, our proposed framework does provide solutions for both mission and motion plans that are reactive. First, we develop a modification of the $L^*$ learning algorithm \cite{angluin1987learning} such that it can be applied for local mission planning even the agent's model is not known {\it a priori}; secondly, by composing safe motion primitives, the designed motion controller can reactively interact with (possibly) uncertain environment and with other agents. For example, collisions with either obstacles or other agents are avoided.
\end{enumerate}

The remainder of this paper is organized as follows. Previous work related to multi-agent coordination and control are briefly reviewed in Section \ref{sec:relatedwork}. After introducing necessary preliminaries in Section \ref{sec:preliminaries}, Section \ref{sec:Problem formulation} presents the formal statement of our problem, along with a motivating example that are used throughout the rest of the paper. Section \ref{sec:Preliminary Results} solves the top-down problem while Section \ref{sec:CoSMoP} solves the bottom-up design problem. Section \ref{sec:conclusion} concludes the paper.

\section{Related Work}\label{sec:relatedwork}
In this paper, we leverage guidance from this relatively broad body of literature to develop a formal framework to solve the mission and motion planning problems of cooperative multi-agent systems. It is worth noting that the proposed framework shows great features of both bottom-up and top-down design methods. Using such a framework, we demonstrate how formal synthesis and verification techniques can facilitate the design of coordination and control protocols for cooperative multi-agent systems.

\subsection{Multi-Agent Systems}
The increasing interest in improving the expressiveness of mission and/or motion planning specifications draws our attention to specifying desired multi-agent behavior in the form of {\it formal languages}, such as regular languages and temporal logics including linear temporal logic (LTL) and computation tree logic (CTL) \cite{baier2008principles}, which provide formal means of specifying high-level performance objectives due to their expressive power. A common two-layered architecture is usually deployed in the synthesis problems of the formal specifications \cite{tabuada2006linear}\cite{kloetzer2008fully}. Based on constructing appropriate finite-state abstractions of not only the underlying dynamical system, but the working environment as well, a control strategy \cite{bloem2012synthesis}, usually represented by a finite state automaton, is synthesized for the satisfaction of the high-level specifications by using formal methods, including model checking \cite{baier2008principles}, supervisory control theory \cite{ramadge1987supervisory}\cite{cassandras2009introduction} and reactive synthesis \cite{bloem2012synthesis}. This synthesis procedure leads to a hierarchical control structure with a discrete planner that is responsible for the high-level, discrete plan and a corresponding low-level continuous controller. Simulations and bisimulation relations are established \cite{baier2008principles} as a proof that the continuous execution of the low-level controller preserves the correctness of the high-level discrete plans \cite{fainekos2009temporal}\cite{kress2009temporal}.

\subsection{Bottom-up Synthesis}
One of the most highlighted bottom-up methods in literature can be categorized as the {\it behavior-based} \cite{arkin1998behavior} approaches, which coordinate multiple agents by composing pre-defined behaviors or distributed learning algorithms from artificial intelligence \cite{ferber1999multi}. It turns out, however, that much of this behavior-based work possesses empirical features that leads to a trial-and-error design process, and therefore lacks guarantees of high-level performance objectives. Recent studies \cite{lyons2012designing}\cite{lyons2015performance} have accounted for performance verification of behavior-based schemes; nevertheless, the contribution are mainly made to single-agent cases. To accomplish high-level tasks of cooperative multi-agent systems, many attempts have been made in the context of bottom-up design. Filippidis et al. \cite{filippidis2012decentralized} proposed a decentralized control architecture of multi-agent systems to address local linear temporal logic (LTL) \cite{baier2008principles} specifications while obeying inter-agent communication constraints; however, the agents therein did not impose any constraints on other agents' behavior. Guo and Dimarogonas \cite{guo2015multi} considered the synthesis of motion plans associated with each agent to fulfill corresponding local LTL specifications by developing a partially decentralized solution which formed clusters of dependent agents such that all individual tasks can be finished in an orderly manner. To overcome the computational issues, the results were further extended in \cite{tumova2016multi} by involving receding horizon planning techniques.

\subsection{Top-down Synthesis}
Karimadini and Lin \cite{karimadini2011guaranteed} studied task decomposition problems of cooperative multi-agent systems, and necessary and sufficient conditions were derived under which the global tasks can be retrieved by the assigned local specifications in the sense of bisimulation \cite{baier2008principles}. Task decomposition from a computationally tractable fragment of computation tree logic (CTL) specifications were also investigated by Partovi and Lin \cite{partovi2014assume}. Following a top-down architecture, Kloetzer and Belta \cite{kloetzer2010automatic} solved the multi-agent coordination problem from a global LTL specification, by model checking the composed behavior of all agents in a centralized manner; the results were extended in \cite{ulusoy2013optimality}, in which optimality and robustness properties of the synthesized motion plans were taken into consideration. ``Trace-closed" regular specifications were investigated in \cite{chen2012formal}\cite{ding2011automatic} to automatically deploy cooperative multi-agent teams. Karaman and Frazzoli \cite{karaman2011linear} addressed the mission planning and routing problems for multiple uninhabited aerial vehicles (UAV), in which the given LTL specifications can be systematically converted a set of constraints suitable to a mixed-integer linear programming (MILP) formulation.

Furthermore, even though powerful model checking tools have been exploited \cite{belta2007symbolic}\cite{bhatia2010sampling} to synthesize control protocols for formal specifications, these approaches generate open-loop strategies and cannot handle reactive specifications; furthermore, those synthesis methods which work for reactive control protocols \cite{kress2009temporal} are severely limited by their high computational complexity. To mitigate this problem, Wongpiromsarn et al. \cite{wongpiromsarn2012receding} employed a receding horizon process where a controller only repeatedly worked out a plan for a short time horizon ahead of the current status. Nevertheless, the proposed results have difficulty handling cooperative tasks for multi-agent systems that involved close inter-agent cooperation.

\subsection{Symbolic Motion Planning}
Control theory has been widely involved to develop performance-guaranteed solutions of planning problems. The classical {\it reach-avoid} planning and {\it point-to-point} motion planning algorithms \cite{chosetprinciples}\cite{lavalle2006planning} aim to steer an intelligent agent from a given initial position to some desirable final configuration while avoiding the collision with any obstacles along the way by utilizing various graph search techniques. Nevertheless, exact solutions to this problem are generally intractable, and various efforts have been devoted to efficiently overcoming the computational burden \cite{bhatia2010sampling}\cite{karaman2011sampling}. It turns out that extension of single-agent planning algorithms to multi-agent cases can be non-trivial, whereas many attempts have been made to achieve different multi-agent coordination and control purposes, such as consensus \cite{olfati2004consensus}\cite{ren2008consensus}, flocking \cite{olfati2006flocking}\cite{zavlanos2009hybrid}, rendezvous \cite{dimarogonas2007rendezvous} and formation control \cite{fax2004information} of multi-agent systems. Fulfillment of these coordination goals is ensured by control theoretical analysis and deductive verification, including Lyapunov stability \cite{olfati2004consensus}\cite{olfati2006flocking} analysis, barrier certificates \cite{panagou2016distributed}, differential dynamic logic \cite{platzer2010logical}, and game theory \cite{semsar2009multi}\cite{marden2009cooperative}. However, these traditional planning and coordination approaches guarantee the steady-state performance of the underlying multi-agent systems, whereas satisfaction of more complex and temporal specifications is not considered.

\subsection{Integrated Task and Motion Planning}
Traditionally, the high-level task planner for mobile robots sits on top of the motion planner \cite{latombe2012robot}. The task planner sees the world as abstracted symbols and ignores details in geometric or physical constraints, which may cause infeasibility in the motion planning. Therefore, a recent trend is towards an Integrated Task and Motion Planning (ITMP). Earlier efforts in ITMP, such as Asymov \cite{cambon2003overview} and SMAP  \cite{plaku2010sampling}, were still based on abstractions of the working environment and used a symbolic planner to provide a heuristic guidance to the motion planner. Recent work, such as \cite{littlefield2014extensible} and \cite{dornhege2012semantic},  introduced  a ``semantic attachment,"  i.e. a predicate that is solved by a motion planner, to the symbolic planner.  An overview of the recent developments in the symbolic motion planning can be found in \cite{lin2014mission}, where the task planning problem is reduced to model checking. Since these methods are based on abstracted symbolic models of the environments, it is a common assumption that the working environment is known or static and the robot is the only moving object (or the robot itself carries  other movable objects). However, in practice, a robot often shares its workspace with others robots or even humans, and the environment often changes over time in a way that is hard to predict.

\section{Preliminaries}\label{sec:preliminaries}
In this section, we introduce the basic concepts and notations that are used throughout this paper to describe cooperative multi-agent systems and their desired properties.
\subsection{Regular Languages}
For a finite set $\Sigma$, we let $2^\Sigma$ and $|\Sigma|$ denote the powerset and the cardinality of $\Sigma$, respectively; furthermore, let $\Sigma^*$, $\Sigma^+$ and $\Sigma^\omega$ denote the set of finite, non-empty finite and infinite sequences that consist of elements from $\Sigma$. A finite sequence $w$ composing of elements in $\Sigma$, i.e., $w=w(0)w(1)\ldots w(n)$, is called a {\it word} over $\Sigma$. The length of a word $w\in \Sigma^*$ is denoted by $|w|$. For two finite words $w_1$ and $w_2$, let $w_1w_2$ denote the word obtained by concatenating $w_1$ and $w_2$. A finite word $s\in \Sigma^*$ is said to be a {\it prefix} of another word $t\Sigma^+$, written as $s\le t$, if there exists a word $u$ such that $t=su$.

Given a finite event set $\Sigma$, a subset of words in $\Sigma^*$ is called a (finite) {\it language} over $\Sigma$. For a language $K\subseteq \Sigma^*$, the set of all {\it prefixes} of words in $K$ is said to be the {\it prefix-closure} of $K$, denoted by $\overline{K}$, that is, $\overline{K}=\{s\in\Sigma^*|\exists t\in\Sigma^*: st\in K\}$, where $st$ denotes the concatenation of two words $s$ and $t$. $K$ is said to be {\it prefix-closed} if $\overline{K}=K$. In practice, we use {\it deterministic finite automata} to recognize languages.

\begin{defn}[Deterministic Finite Automaton]
	A deterministic finite automaton (DFA) is a 5-tuple
	$$G=(Q,\Sigma,q_0,\delta,Q_m),$$
	where $Q$ is a finite set of states, $\Sigma$ is a finite set (alphabet) of events, $q_0\in Q$ is an initial state, $\delta: Q\times \Sigma \to Q$ is a partial transition function and $Q_m\subseteq Q$ is the set of the marked (accepting) states.
\end{defn}

The transition function $\delta$ can be generalized to $\delta: Q\times \Sigma^* \to Q$ in the usual manner \cite{cassandras2008introduction}. The language generated by $G$ is defined as $L(G):=\left\{s\in \Sigma^* \vert \delta(q_0,s)\mbox{ is defined.}\right\}$; while $L_m(G)=\left\{s\in \Sigma^* \vert s\in L(G), \delta(q_0,s)\in Q_m\right\}$ stands for the language that is marked by $G$. The language that is accepted by a DFA is called a {\it regular language}. We focus our study on regular languages in the sequel.

For a non-empty subset $\Sigma'\subseteq \Sigma$ and a word $s$ over $\Sigma$, we use the ``natural projection" to form a word $s'$ over $\Sigma'$ from $s$ by eliminating all the events in $s$ that does not belong to $\Sigma'$. Formally, we have

\begin{defn}[Natural Projection]
	For a non-empty subset $\Sigma'\subseteq \Sigma$, the natural projection $P: \Sigma^*\to \Sigma'^*$ is inductively defined as
	$$P(\epsilon)=\epsilon$$
	$$
	\forall s\in \Sigma^*, \sigma\in \Sigma, P(s\sigma)=
	\begin{cases}
	P(s)\sigma,  & \mbox{if } \sigma\in \Sigma', \\
	P(s), &\mbox{otherwise.}
	\end{cases}
	$$
	The set-valued {\it inverse projection} $P^{-1}:2^{\Sigma'^*} \to 2^{\Sigma^*}$ is of $P$ defined as $P^{-1}(s)=\left\{t\in \Sigma^*: P(t)=P(s)\right\}$.
\end{defn}

Given a family of event sets $\{\Sigma_i\}$, $i=1,2, \ldots, N$, with their union $\Sigma=\bigcup_{i=1}^N \Sigma_i$, we let $P_i$ denote the natural projection from $\Sigma$ to $\Sigma_i$. For a finite set of regular languages $L_i\subseteq \Sigma_i^*$, $i=1,2, \ldots, N$, the {\it synchronous product} of $\{L_i\}$, denoted by $\vert\vert_{i=1}^n L_i$, is defined as follows.

\begin{defn}[Synchronous Product]\label{product}
	\cite{willner1991supervisory} For a finite set of regular languages $L_i\subseteq \Sigma_i^*$, $i=1,2, \ldots, N$,
	\begin{equation}
	\vert\vert_{i=1}^n L_i=\{t\in \Sigma^*\vert \forall i: P_i(t)\in L_i\}.
	\end{equation}
	Equivalently, $\vert\vert_{i=1}^n L_i=\bigcap_{i=1}^n P_i^{-1}(L_i).$
\end{defn}

\subsection{Differential Dynamic Logic}

The Differential Dynamic Logic $d\mathcal{L}$ verifies a symbolic hybrid system model, and, thus, can assist in verifying and finding symbolic parameters constraints. Most of the time, this turns into an undecidable problem for model checking \cite{platzer2010logical}. Yet, the iteration between the discrete and continuous dynamics is nontrivial and leads to nonlinear parameter constraints and nonlinearities in the dynamics. Hence, the model checking approach must rely on approximations. On the other hand, the $d\mathcal{L}$ uses a deductive verification approach to handling infinite states, it does not rely on finite-state abstractions or approximations, and it can handle those nonlinear constraints.

The hybrid systems are embedded to the d$\mathcal{L}$ as hybrid programs, a compositional program notation for hybrid systems.

\begin{defn}[Hybrid Program]\label{def:hybridprogram}
	A hybrid program \cite{platzer2010logical} ($\alpha$ and $\beta$) is defined as:
	\begin{equation*}
	\alpha, \beta ::= \begin{cases}
	x_1 := \theta_1,...,x_n:=\theta_n \mid  ?\chi \mid \alpha ; \beta \mid \alpha \cup \beta \mid \alpha^* \mid \\
	x_1^{\prime} := \theta_1,...,x_n^{\prime}:=\theta_n \& \chi			
	\end{cases}
	\end{equation*}
	where:
	\begin{itemize}
		\item $x$ is a state variable and $\theta$ a first-order logic term.
		\item $\chi$ is a first-order formula.
		\item $x_1 := \theta_1,...,x_n:=\theta_n$ are discrete jumps, i.e. instantaneous assignments of values to state variables.
		\item $x_1^{\prime} := \theta_1,...,x_n^{\prime}:=\theta_n \& \chi$ is a differential equation system that represents the continuous variation in system dynamics. $x_i^{\prime} := \theta_i$ is the time derivative of state variable $x_i$, and $\& \chi$ is the evolution domain.
		\item $?\chi$ tests a first-order logic at current state.
		\item $\alpha ; \beta$ is a sequential composition, i.e. the hybrid program $\beta$ will start after $\alpha$ finishes.
		\item $\alpha \cup \beta$ is a nondeterministic choice.
		\item $\alpha^*$ is a nondeterministic repetition, which means that $\alpha$ will repeat for finite times.
	\end{itemize}
\end{defn}

Thus, we can define the $d\mathcal{L}$ formula, which is a first-order dynamic logic over the reals for hybrid programs.

\begin{defn}[$d\mathcal{L}$ formulas]
	A $d\mathcal{L}$ formula \cite{platzer2010logical} ($\phi$ and $\psi$) is defined as:
	\begin{equation*}
	\phi, \psi ::= \chi \mid \neg \phi \mid \phi \wedge \psi \mid \forall x \phi \mid \exists x \phi \mid [\alpha] \phi \mid \langle \alpha \rangle \phi
	\end{equation*}
	where:
	\begin{itemize}
		\item $[\alpha] \phi$ holds true if $\phi$ is true after all runs of $\alpha$.
		\item $\langle \alpha \rangle \phi$ holds true if $\phi$ is true after at least one runs of $\alpha$.
	\end{itemize}
\end{defn}

$d\mathcal{L}$ uses a compositional verification technique that permits the reduction of a complex hybrid system into several subsystems \cite{platzer2010logical}. This technique divides a system $\psi \rightarrow [\alpha] \phi$ in an equivalent formula $\psi_1 \rightarrow [\alpha_1] \phi_1 \wedge \psi_2 \rightarrow [\alpha_2] \phi_2$, where each $\psi_i \rightarrow [\alpha_i] \phi_i$ can be proven separately. In our approaches we use this technique backwards, we prove a set of $d\mathcal{L}$ formulas $\psi_i \rightarrow [\alpha_i] \phi_i$, where each one is the $i^{th}$ motion primitive model, and we use the SMT to compose an equivalent $\psi \rightarrow [\alpha] \phi$ that satisfies a mission task. Therefore, the synthesized hybrid system performance is formally proven.

%
%

\subsection{Counter Linear Temporal Logic Over Constraint System}\label{sec:CLTLBD}

We express the specification of an autonomous mobile robot using Counter Linear Temporal Logic Over Constraint System CLTLB($\mathcal{D}$) defined in \cite{bersani2010bounded}. This language is interpreted over Boolean terms $p \in AP$ or arithmetic constraints $R \in \mathcal{R}$ belong to a general constraint system $\mathcal{D}$, where $AP$ is a set of atomic propositions and $\mathcal{R}$ is a set of arithmetic constraints. Thus, the semantics of a CLTLB($\mathcal{D}$) formula is given in terms of interpretations of a finite alphabet $\Sigma \in \{AP, \mathcal{R}\}$ on finite traces over a finite sequence $\rho$ of consecutive instants of time with length $K$, meaning that $\rho(k)$ is the interpretation of $\Sigma$ at instant of time $k \in \mathcal{N}_{\rho}, \mathcal{N}_{\rho} = \{0,...,K\}$. Moreover, the arithmetic terms of an arithmetic constraint $R \in \mathcal{R}$ are variables $x$ over a domain $D \in \{\mathbb{Z}, \mathbb{R}\}$ valuated at instants $i$ and, thus, are called arithmetic temporal terms \textit{a.t.t.},

\begin{defn}[Arithmetic Temporal Term]
	A CLTLB($\mathcal{D}$) arithmetic temporal term (\textit{a.t.t.}) $\varphi$ is defined as:
	\begin{equation*}
	\varphi ::= x \mid \bigcirc \varphi \mid \bigcirc^{-1} \varphi
	\end{equation*}
	where $\bigcirc$ and $\bigcirc^{-1}$ stands for next and previous operator.
\end{defn}

Therefore, a CLTLB($\mathcal{D}$) formula is a LTL formula over the \textit{a.t.t.} defined as below.

\begin{defn}[Formula]
	A CLTLB($\mathcal{D}$) formula ($\phi$, $\phi_1$ and $\phi_2$) is defined as,
	\begin{equation*}
	\phi, \phi_1, \phi_2 ::=
	\begin{cases}
	p \mid R(\varphi_1,...,\varphi_n) \mid \neg \phi \mid \phi_1 \wedge \phi_2 \mid \\
	\bigcirc \phi  \mid \bigcirc^{-1} \phi \mid \phi_1 \mathbf{U} \phi_2 \mid \phi_1 \mathbf{S} \phi_2
	\end{cases}
	\end{equation*}
	where,
	\begin{itemize}
		\item $p \in AP$ is a atomic proposition, and $R \in \mathcal{R}$ is a relation over the \textit{a.t.t.} such as, for this work, we limit it to linear equalities or inequalities, i.e. $R(\varphi_1,...,\varphi_n) \equiv \sum_{i=1}^{n} c_i \cdot \varphi_i \# c_0$, where $\# \equiv \langle =, <, \leq, >, \geq \rangle$ and  $c_i, \varphi_i \in D$.
		\item $\bigcirc$, $\bigcirc^{-1}$, $\mathbf{U}$ and $\mathbf{S}$ stands for usual next, previous, until and since operators on finite traces, respectively.
	\end{itemize}
\end{defn}

Based on this grammar, it can also use others common abbreviations, including:

\begin{itemize}
	\item Standard boolean, such as $true$, $false$, $\vee$ and $\rightarrow$.
	\item $\Diamond \phi$ that stands for $true \mathbf{U} \phi$, and it means that $\phi$ eventually holds before the last instant (included).
	\item $\square \phi$ that stands for $\neg \Diamond \neg \phi$, and it means that $\phi$ always holds until the last instant.
	\item $Last [\phi]$ that stands for $\Diamond (\neg \bigcirc true) \wedge \phi$, where $\neg \bigcirc true$ on finite trace is only $true$ at last instant. Thus, it means that $\phi$ is true at the last instant of the sequence $\rho$.
\end{itemize}

A CLTLB($\mathcal{D}$) formula is verified in a Bounded Satisfiability Checking (BSC) \cite{pradella2013bounded}. Hence, it is interpreted on a finite sequence $\rho$ with length $K$. Therefore, $\rho(k) \vDash p$ means that $p$ holds true in the sequence $\rho$ at instant $k$ ($p \vdash \rho(k)$).

\begin{defn}[Semantics]
	The semantics of a CLTLB($\mathcal{D}$) formula $\phi$ at an instant $k \in \mathcal{N}_{\rho}$ is as follow:
	
	\begin{itemize}
		\item $\rho(k) \vDash p \Longleftrightarrow p \vdash \rho(k)$.
		\item $\rho(k) \vDash R(\varphi_1,..., \varphi_n)  \Longleftrightarrow R(\varphi_1,..., \varphi_n) \vdash \rho(k)$.
		\item $\rho(k) \vDash \neg \phi \Longleftrightarrow \rho(k) \nvDash \phi$.
		\item $\rho(k) \vDash \phi_1 \wedge \phi_2 \Longleftrightarrow \rho(k)  \vDash \phi_1 \wedge \rho(k)  \vDash \phi_2$.
		\item $\rho(k) \vDash \bigcirc \phi \Longleftrightarrow \rho(k+1)  \vDash \phi$.
		\item $\rho(k) \vDash \bigcirc^{-1} \phi \Longleftrightarrow \rho(k-1)  \vDash \phi$.
		\item $\rho(k) \vDash \phi_1 \mathbf{U} \phi_2 \Longleftrightarrow \begin{cases}
		\exists i \in [k,K]: \rho(i)  \vDash \phi_2 \wedge \\
		\forall j \in [k,i-1]: \rho(j) \vDash \phi_1
		\end{cases}$.
		\item $\rho(k) \vDash \phi_1 \mathbf{S} \phi_2 \Longleftrightarrow \begin{cases}
		\exists i \in [0,k]: \rho(i)  \vDash \phi_2 \wedge \\
		\forall j \in [i+1,k]: \rho(j) \vDash \phi_1
		\end{cases}$.
	\end{itemize}
\end{defn}

\section{Problem formulation}\label{sec:Problem formulation}
\subsection{A Motivating Example}
\begin{figure}
	\centering
	\includegraphics[width=2.5in]{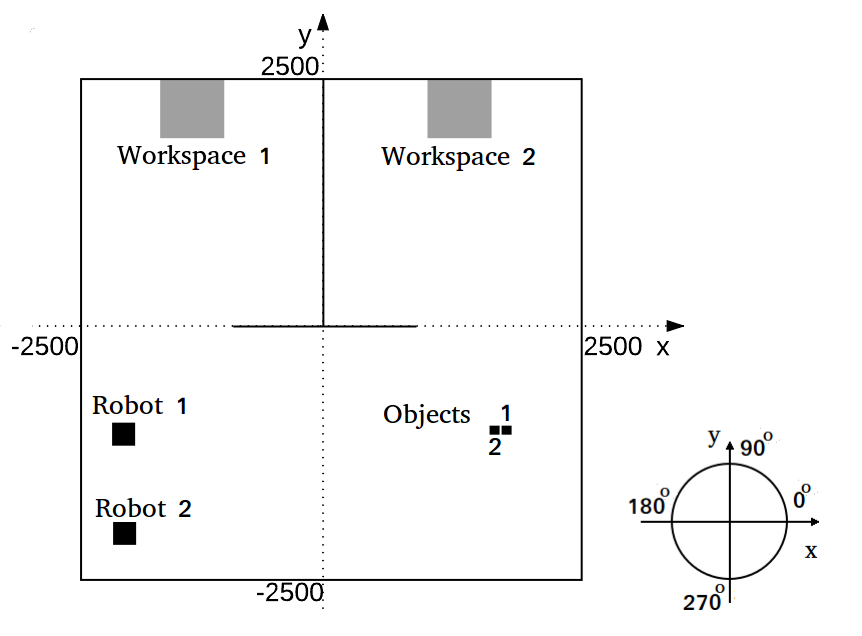}
	\caption{Warehouse layout}
	\label{fig:example01}
\end{figure}

As a motivating example, let us consider a cooperative MRS with $N$ robots in an automated warehouse as shown in Fig. \ref{fig:example01}. The global mission is to deploy the robots to move newly arrived goods to respectively designated workspaces. Additionally, the robots are Pioneer P3-DX  robots\footnote{http://www.mobilerobots.com/ResearchRobots/PioneerP3DX.aspx, retrieved 05-18-2016.} which is fully programmable and includes a dedicated motion controller with encoder feedback. Moreover, this robot can be simulated with the MobileSim\footnote{http://www.mobilerobots.com/Software/MobileSim.aspx, retrieved 05-18-2016.}. This application permits to simulate all current and legacy models of MobileRobots/ActivMedia mobile robots such as Pioneer 3 DX and AT. Moreover, full source code is available under the GPL for understanding the simulation implementation, customizing and improving it.

Furthermore, the Pioneer P3-DX robot has a software developing kit called Pioneer SDK\footnote{http://www.mobilerobots.com/Software.aspx, retrieved 05-18-2016.} which allows developing its control system in custom C++ applications with third-part libraries such as an SMT solver. Particularly, the examples presented in this paper are implemented using two libraries from this kit: ARIA\footnote{http://www.mobilerobots.com/Software/ARIA.aspx, retrieved 05-18-2016.} and ARNL\footnote{http://www.mobilerobots.com/Software/NavigationSoftware.aspx, retrieved 05-18-2016.}. The ARIA brings an interface to control and to receive data from MobileSim accessible via a TCP port and is the foundation for all other software libraries in the SDK such as the ARNL. Moreover, the ARNL Navigation library\footnote{http://www.mobilerobots.com/Software/NavigationSoftware.aspx, retrieved 05-18-2016.} provides a MobileRobots' proprietary navigation technology that is reliable, high quality and highly configurable and implement an intelligent navigation and positioning capabilities to this robot. Different localization (positioning) methods are available for various sensors such as LIDAR, Sonar, and GPS. Furthermore, commands can be sent to a custom application implementing those libraries by using a graphical interface called MobileEye\footnote{http://www.mobilerobots.com/Software/MobileEyes.aspx, retrieved 05-18-2016.} which shows the sensor readings and trajectories. Hence, each robot dynamics is simulated in MobileSim, and each controller is implemented in a C++ custom application that both run on Linux Computers. These computers are connected through Ethernet, and each robot controller connects via a TCP port to MobileSim and other robot neighbors. Therefore, all examples presented in this paper can be implemented in a custom C++ application using both the Pioneer SDK and the SMT solver (e.g. Z3).

This article illustrates the control system design using a simple example shown in the Fig. \ref{fig:example01}. Denote $\mathcal{N}_{\mathcal{A}}=\{1,...,N\}$, we initially assume $N=2$ and all the robots $R_i$, $i\in\mathcal{N}_{\mathcal{A}}$ have the identical communication, localization and actuation capabilities. Our design framework can be extended to involve $N>2$ robots with different capabilities and other scenarios like search and rescue as well, and it will be presented an example with $N=10$.

This robot team may share its workspace with humans and deal with unexpected obstacles such as a box that falls from a shelf. Some goods must be moved first before the others can be picked up, some maybe quite heavy and require two robots to move; therefore coordination between robots is needed for the safety as well as the accomplishment of the global task.

\subsection{Cooperative Mission Planning Problem}
Motivated by the fact that the accomplishment of missions among cooperative multi-agent systems shows strong event-driven features, we characterize the mission planning problem within the discrete-event system (DES) formalism \cite{cassandras2008introduction}. For a cooperative multi-agent system that consists of $N$ interacting agents, let $\Sigma_{MI}^i$ denote the set of missions that can be accomplished by the $i$-th agent, $i\in\mathcal{N}_{\mathcal{A}}$. In practical mission planning problems, events in $\Sigma_{MI}^i$ shall represent the sensing, and actuating capabilities of the underlying agent; and execution of an event $\sigma_{MI}^i$ indicates that the $i$-th agent may accomplish a certain action. The ``global" missions are then captured by the union of mission capabilities of all agents, i.e., $\Sigma_{MI}=\bigcup_{i\in\mathcal{N}_{\mathcal{A}}} \Sigma_{MI}^i$. For the clarity of presentation, we assume that the mission transition diagram of each agent $R_i$ is given by a prefix-closed regular language $K_{MI}^i\subseteq \Sigma_{MI}^{i*}$.

The mission alphabet $\Sigma^i_{MI}$, $i\in\mathcal{N}_{\mathcal{A}}$ of the motivating example are listed in Table \ref{t:table} with an explanation of the corresponding service and mission capabilities.
\begin{table}[H]
	\begin{center}
		\caption{$\Sigma^i_{MI}$} 
		\begin{tabular}{ll} \toprule
			\multicolumn{1}{l}{Event} &  {Explanation} \\ \midrule
			$R_ipO_j$ & Robot $R_i$ picks up object $O_j$. \\ 
			$R_idO_jaW_k$ & Robot $R_i$ drops off $O_j$ at workspace $k$.\\
			$r_i$ & Robot $R_i$ returns to its original position.\\
			$O_jAway$ & $O_j$ is moved away
			\\\bottomrule
			$i,j,k=1,2$
		\end{tabular} \label{t:table}
	\end{center}
	\vspace{-5 mm}
\end{table}

Inter-agent communication for the purpose of multi-agent coordination are considered at this point by imposing extra constraints on events shared by more than one agent.  For each agent $R_i$, $i\in\mathcal{N}_{\mathcal{A}}$, we associate a pair of request and response {\it communication events}, respectively as follows:
$$\Sigma^{req,i}=\{?\sigma|(\exists j\ne i)\sigma\in(\Sigma_{MI}^i\cap\Sigma_{MI}^j)\},$$
and
$$\Sigma^{res,i}=\{!\sigma|(\exists j\ne i)\sigma\in(\Sigma_{MI}^i\cap\Sigma_{MI}^j)\},$$
where a request event indicates that the underlying agent sends a message through the communication channel, and a response event indicates a message reception. In the warehouse example, $O_jAway$ is a communication event where $?O_jAway$ denotes a request event that some robot wants the $O_j$ to be moved away. $!O_jAway$ denotes a response event that some robot moves $O_j$ away and notifies the robot who made the request. 

\begin{figure}
	\centering	
	\begin{tikzpicture}[shorten >=1pt,node distance=2cm,on grid,auto, bend angle=20, thick,scale=0.55, every node/.style={transform shape}]
	\node[state,initial] (s_0)   {};
	\node[state] (s_1) [right=of s_0] {};
	\node[state] (s_2) [right=of s_1] {};
	\node[state] (s_3) [right=of s_2] {};
	\node[state] (s_4) [below=of s_0] {};
	\node[state] (s_5) [right=of s_4] {};
	\node[state] (s_6) [right=of s_5] {};
	\node[state] (s_7) [right=of s_6] {};
	\node[state] (s_8) [below=of s_4] {};
	\node[state] (s_9) [right=of s_8] {};
	\node[state] (s_10) [right=of s_9] {};
	\node[state] (s_11) [right=of s_10] {};
	\node[state] (s_12) [below=of s_8] {};
	\node[state] (s_13) [right=of s_12] {};
	\node[state] (s_14) [right=of s_13] {};
	\node[state] (s_15) [right=of s_14] {};
	\path[->]
	(s_0) edge node [pos=0.5, sloped, above=0.5]{$R_1pO_1$} (s_1)
	(s_0) edge node [pos=0.5, sloped, below=0.5]{$R_2pO_2$} (s_4)
	(s_1) edge node [pos=0.5, sloped, above=0.5]{$R_1dO_1aW_1$} (s_2)
	(s_1) edge node [pos=0.5, sloped, above=0.5]{} (s_5)
	(s_2) edge node [pos=0.5, sloped, above=0.5]{$r_1$} (s_3)
	(s_2) edge node [pos=0.5, sloped, above=0.5]{} (s_6)
	(s_3) edge node [pos=0.5, sloped, below=0.5]{$R_2pO_2$} (s_7)
	(s_4) edge node [pos=0.5, sloped, below=0.5 ]{$R_2dO_2aW_2$} (s_8)
	(s_4) edge node [pos=0.5, sloped, above]{} (s_5)
	(s_5) edge node [pos=0.5, sloped, above]{} (s_6)
	(s_5) edge node [pos=0.5, sloped, above]{} (s_9)
	(s_6) edge node [pos=0.5, sloped, above]{} (s_7)
	(s_6) edge node [pos=0.5, sloped, above]{} (s_10)
	(s_7) edge node [pos=0.5, sloped, below=0.5]{$R_2pO_2$} (s_11)
	(s_8) edge node [pos=0.5, sloped, below=0.5 ]{$r_2$} (s_12)
	(s_8) edge node [pos=0.5, sloped, above]{} (s_9)
	(s_9) edge node [pos=0.5, sloped, above]{} (s_10)
	(s_9) edge node [pos=0.5, sloped, above]{} (s_13)
	(s_10) edge node [pos=0.5, sloped, above]{} (s_11)
	(s_10) edge node [pos=0.5, sloped, above]{} (s_14)
	(s_11) edge node [pos=0.5, sloped, below=0.5]{$r_2$} (s_15)
	(s_12) edge node [pos=0.5, sloped, below=0.5]{$R_1pO_1$} (s_13)
	(s_13) edge node [pos=0.5, sloped, below=0.5]{$R_1dO_1aW_1$} (s_14)
	(s_14) edge node [pos=0.5, sloped, below=0.5]{$r_1$} (s_15);
	\end{tikzpicture}
	\caption{Global specification}
	\label{fig:spec}
\end{figure}
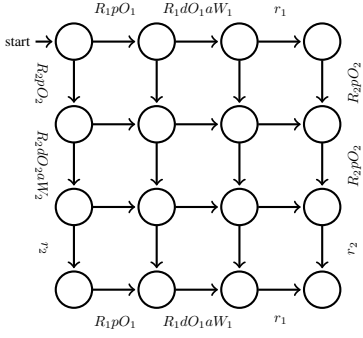	
The team task $K_{MI}\subseteq \Sigma_{MI}^*$ is given in the form of a prefix-closed regular language associated with its DFA representation. The design objective of the mission planning is to decompose the global mission into local tasks $K^i_{MI}$, $i\in\mathcal{N_A}$, such that $\vert\vert_{i\in\mathcal{N}_{\mathcal{A}}} K^i_{MI} \models K_{MI}$,  i.e., $\left(\vert\vert_{i\in\mathcal{N}_{\mathcal{A}}} K^i_{MI}\right) \subseteq K_{MI}$. That is, the collective team behavior should not exceed the global mission. In summary, the top-down design objective is to solve the following distributed cooperative tasking problem.


\begin{problem}[Cooperative Mission Planning]\label{problem:p1}
	Given a non-empty and prefix-closed global mission $K_{MI}$ and $\Sigma_{MI}$, local mission sets $\Sigma^i_{MI},i\in\mathcal{N}_{\mathcal{A}}$ of each robot, systematically find locally feasible mission plans $K^i_{MI}$ for each robot such that $\vert\vert_{i\in\mathcal{N}_{\mathcal{A}}} K^i_{MI} \models K_{MI}$.	
\end{problem}

The team mission for the automated warehouse example is as shown in Fig. \ref{fig:spec}. All the horizontal events of the same column and all the vertical events of the same row are identical.
\subsection{Integrated task and motion planning}
Given the local mission plan $K^i_{MI}$ for each robot $R_i$, the underlying integrated task and motion planning problem is to implement the task with safety guarantees.

The description of the scenario environment is essential for the integrated task and motion planning problem. Hence, we first define the scene description which provides the basic information of the robot workspace. Since the Pioneer P3-DX robot is a ground vehicle, its workspace can be specified in 2D.
\begin{defn}[Scene Description] \label{def:scene}
	\rm Scene description is a tuple $\mathcal{M} = \langle \mathcal{O}, \mathcal{A}, \mathcal{B} \rangle$:
	\begin{itemize}
		\item Obstacles $\mathcal{O}$: a set of polygon obstacles described by line segments $o_j, j \in \mathcal{N}_{\mathcal{O}}$ specified by two points $o_i = \langle (x_i, y_i), (x_f,  y_f) \rangle$, where $\mathcal{N}_{\mathcal{O}} = \{1,...,|\mathcal{O}| \}$;
		\item Agents $\mathcal{A}$: a set of robots $R_i \in \mathcal{A} : R_i = \langle l, q_{r,0} \rangle, i\in \mathcal{N}_{\mathcal{A}}$ which are represented as a square described by their length $l$ and their initial state $q_{r,0}$.
		\item Objects $\mathcal{B}$: a set of movable objects $b_j = \langle l, q_{b,0} \rangle, j \in \mathcal{N}_{\mathcal{B}}$ which are specified as an square described by their length $b_i.l$ and their initial state $q_{b,0}$, where $\mathcal{N}_{\mathcal{B}} = \{1,...,|\mathcal{B}|\}$.
	\end{itemize}
\end{defn}

The states variables of the robots and objects are defined over instants of time indicating the execution ending events of the primitives. Those instants of time are defined by $k \in \mathbb{Z}_{\geq 0}$, as defined in Sec.~\ref{sec:CLTLBD}, and it denotes the time instant that the $k$th action has been taken. Thus, we denote the robot $R_i$ state variables as $q_{r}^{i}, i\in\mathcal{N}_\mathcal{A} : q_{r}^{i} = \langle x, y, \alpha \rangle$ which represents the robot pose, where $x,y \in \mathbb{Z}$ specify the position in $mm$ and $\alpha \in \mathbb{R}$ is angle in degrees. Hence, a CLTLB($\mathcal{D}$) formula $\Box (q_{r}^{i}.x = \bigcirc q_{r}^{i}.x)$, for example, means that the robot state variable $x$ value at instant $k$ should be always equal to the value of $x$ at $k+1$. Correspondingly, the object state variables are expressed as $q_{b}^{j}, j \in \mathcal{N}_{\mathcal{B}} : q_{b}^{j} = \langle x, y, p, a \rangle$ which describes its 2D position $\langle x, y \rangle: x, y \in \mathbb{Z}$, and $p$ and $a$ are Boolean propositions that $p$ holds true when the robot is carrying this object, and $a$ holds true when another robot is taking this object away from its initial position. Next, we define a scene description for the particular scenario as shown in Fig. \ref{fig:example01}

\begin{example}\label{ex:example01}
	\rm The Fig. \ref{fig:example01} a scene of an automated warehouse which two robots must drop two objects off in two different workspaces. Note that the origin $(0,0)$ is at the center of the workspace. The robots are represented as black filled squares with side length $400mm$ which start at bottom left of this warehouse, i.e. $\mathcal{A} = [\langle 400, (-2000, -1000, 0.0) \rangle, \langle 400, (-2000,-2000, 0.0) \rangle]$. The objects are initially at bottom right of the warehouse and are depicted as black filled square too with side length $100mm$, i.e. $\mathcal{B} = [\langle 100, (2000, -1000, false, false) \rangle, \\\langle 100, (1900, -1000, false,  false) \rangle]$. The obstacles refers to the four boundary lines that limits the scene which are formally specified as a set of line segments $[\langle (-2500,-2500), \\(2500,-2500) \rangle, \langle (2500,-2500),(2500,2500) \rangle, \langle (2500,\\2500), (2500,-2500) \rangle, \langle (2500,-2500), (-2500,-2500) \rangle] \\\subset \mathcal{O}$ and to the two walls that separate the workspaces shown as gray squares, i.e. $\big[\langle (0,0), (0,2500) \rangle, \langle (-1000,0), (1000,\\0) \rangle\big] \subset \mathcal{O}$. The challenge in this scene is that the objects are adjacent to each other; therefore, a plan that includes both robots picking them up at same time requires a cooperative behavior.\qed
\end{example}
\begin{problem}[Reactive Motion Planning]\label{problem:p2}
	Given a team of robots $\mathcal{A}$ and their mission plans $K^i_{MI} : i\in\mathcal{N}_{\mathcal{A}}$, the scene description $\mathcal{M}$, and the trace length $K^i$ for each robot $R_i$, solve an integrated task and motion planning problem by splitting it into three steps. First, design a set of safe motion primitives $\mathcal{P}^i$ for each robot $R_i$ and respective motion primitives specification $\phi_{\mathcal{P}}^i(\mathcal{M})$. A safe motion primitive $\pi^{i,j} \in \mathcal{P}^i : j \in \mathcal{N}_{\mathcal{P}}^i = \{1,...,|\mathcal{P}^i|\}$ for the robot $R_i$ is a certified controller which guarantees a safety property and can be reactive changing its control values based on actual sensor readings. The motion primitives specification $\phi_{\mathcal{P}}^i(\mathcal{M})$ is a CLTLB($\mathcal{D}$) formula which specifies the safe motion primitives by defining constraints for the state variables and the given scene description $\mathcal{M}$. Second, for each robot $R_i$, check if the mission plans $K^i_{MI}$ are satisfiable for the scene $\mathcal{M}$ in a fair environment using the controllers $\mathcal{P}^i$ for each robot $R_i$. An environment is fair when all moving and static obstacles that are not in the scene description do not lead any robot to a deadlock. Third, for all plans $K^i_{MI}$ that are satisfiable, find a trace $s^i$ with length $K^i$ for each robot $R_i$, where $s^i(k) = \langle q_{r}^i(k), \delta^i(k) \rangle$  at instant $k \in \mathcal{N}^i = \{1,...,K^i\}$. $Q_r^i$ is a sequence of assigned values for robot $R_i$ states such as $q_{r}^i(k) \in Q_r^i : q_{r}^i(k) = \langle x, y, \alpha \rangle$ are the values at instant $k$. $Q_{\pi}^i$ is a sequence of assigned primitives such as $\delta^i(k) \in Q_{\pi}^i$ is a motion primitive at instant $k$ that defines to robot $R_i$ what primitive $\pi^{i,j} \in \delta^i(k)$ to take at $q_{r}^i(k-1)$ to go to $q_{r}^i(k)$.
\end{problem}
Note that we are restricted to take at most $K^i$ actions in each mission plan $K^i_{MI}$ and robot $R_i$. The motion controller $\delta^i(k)$ refers to actions that a robot can execute, such as moving to some place, picking up objects and so on. Such actions are designed underlying low-level control law from which the generated trajectories are guaranteed to be safe considering both the environment geometrics and kinematics.

	\section{Top-down design and Task Decomposition}\label{sec:Preliminary Results}

This section concerns with Problem 1 and derives a systematical approach to decompose the global task into feasible local tasks. In our previous work \cite{jinCDC}, a counterexample-guided and learning-based assume-guarantee synthesis framework was proposed. We adopt this framework in the top-down layer in Fig. \ref{fig:overallframework} to automatically learn the local missions $K^i_{MI}$.
\begin{figure}[h]
	\begin{center}
		\centerline{\includegraphics[scale=0.7]{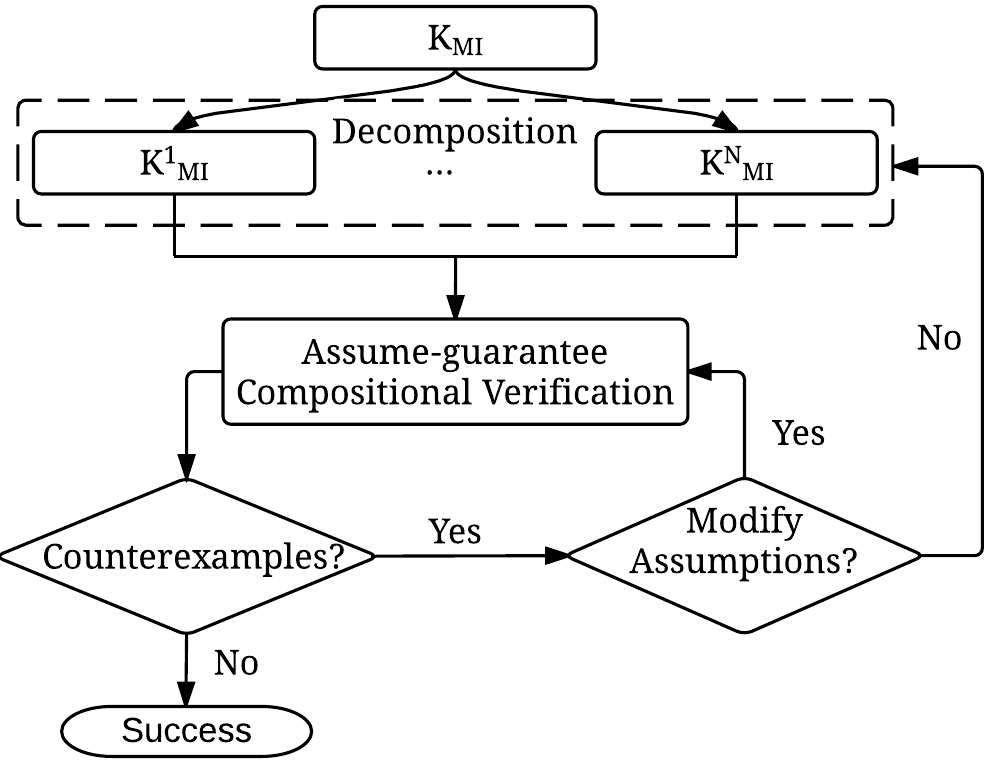}}
		\caption{Learning-based coordination and mission planning framework.}
		\label{fig:caf}
	\end{center}
	\vspace{-7 mm}
\end{figure}

Fig. \ref{fig:caf} shows the flowchart of the automatic task decomposition and coordination framework that solves Problem 1 by executing the following steps iteratively.
\begin{itemize}
	\item {\it Task decomposition} Obtain a prefix-closed and feasible local mission $K^i_{MI}$ for robot $R_i$ from the global mission $K_{MI}$.
	\item {\it Compositional verification} We determine whether or not the collective behaviors of each agent can satisfy the global mission by deploying a compositional verification \cite{jinCDC} procedure with each behavior module being a component DFA that recognizes $K^i_{MI}$. In particular, to mitigate the computational complexity, we adopt an assume-guarantee paradigm for the compositional verification and modify $L^*$ algorithm \cite{angluin1987learning} to automatically learn appropriate assumptions for each agent.
	\item{\it Counterexample-guided synthesis} If the local missions fail to satisfy the global specification jointly, the compositional verification returns a counterexample indicating that all the $K^i_{MI}, i\in \mathcal{N}_{\mathcal{A}}$ share a same illegal trace that violates the global mission. We present such counterexample to re-synthesize the local missions.
\end{itemize}

\begin{figure}
	\centering
	\subfloat[$K^1_{MI}$ for robot $R_1$]{\label{K1}	
		\begin{tikzpicture}[shorten >=1pt,node distance=2.5cm,on grid,auto, bend angle=20, thick,scale=0.55, every node/.style={transform shape}]
		\node[state,initial] (s_0)   {};
		\node[state] (s_1) [right=of s_0] {};
		\node[state] (s_2) [right=of s_1] {};
		\node[state] (s_3) [right=of s_2] {};
		\path[->]
		(s_0) edge node [pos=0.5, sloped, above]{$R_1pO_1$} (s_1)
		(s_1) edge node [pos=0.5, sloped, above]{$R_1dO_1aW_1$} (s_2)
		(s_2) edge node [pos=0.5, sloped, above]{$r_1$} (s_3);
		\end{tikzpicture}}\\
	\subfloat[$K^2_{MI}$ for robot $R_2$]{\label{K2}
		\begin{tikzpicture}[shorten >=1pt,node distance=2.5cm,on grid,auto, bend angle=20, thick,scale=0.55, every node/.style={transform shape}]
		\node[state,initial] (s_0)   {};
		\node[state] (s_1) [right=of s_0] {};
		\node[state] (s_2) [right=of s_1] {};
		\node[state] (s_3) [right=of s_2] {};
		\path[->]
		(s_0) edge node [pos=0.5, sloped, above]{$R_2pO_2$} (s_1)
		(s_1) edge node [pos=0.5, sloped, above]{$R_2dO_2aW_2$} (s_2)
		(s_2) edge node [pos=0.5, sloped, above]{$r_2$} (s_3);
		\end{tikzpicture}}
	\caption{Robots' specifications}
	\label{fig:sup}
\end{figure}
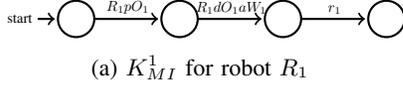
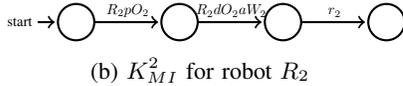

We illustrate the task decomposition using the automated warehouse example in Section II. In the framework shown in Fig.~\ref{fig:caf}, local missions $K^i_{MI}$, $i=1,2$ are obtained by $K^i_{MI}=P_i(K_{MI})$ as shown in Fig.~\ref{fig:sup}, where $P_i$ stands for the {\it natural projection} \cite{cassandras2008introduction} from the global mission set $\Sigma_{MI}$ to the mission set $\Sigma^i_{MI}$ of the $i$-th robot, $i\in \mathcal{N}_A$.  Under the assumption that the global mission is feasible, i.e., $K_{MI}=\overline{K}_{MI}$, we point out that every mission specification $K^i_{MI}$ is locally feasible. 


Given a series of feasible local missions $K^i_{MI}$ for $i=1,2$, the next question is whether or not the fulfillment of all local missions can imply the satisfaction of the global one. This question is addressed by deploying a compositional verification procedure \cite{jinCDC}. Specifically, by setting $K^i_{MI}$ as the $i$-th behavior module, the compositional verification justifies whether or not $M_1\vert\vert M_2\models K_{MI}$ using an assume-guarantee scheme. In the assume-guarantee paradigm for compositional verification, a formula to be checked is a triple $\langle A \rangle M \langle P \rangle $, where $M$ is a module component, $P$ is a property and $A$ is an assumption about $M$'s environment, which can also be represented by a DFA. The formula is true if whenever $M$ is part of a system satisfying $A$, then the system must also guarantee the property $P$, i.e., $\forall E$, $E\vert\vert M \models A$ implies that $E\vert\vert M\models P$. For the warehouse example, we check the achievement of $K_{MI}$ by following an asymmetric proof rule.

\begin{figure}
	\centering
	\begin{tikzpicture}[shorten >=1pt,node distance=2.5cm,on grid,auto, bend angle=20, thick,scale=0.55, every node/.style={transform shape}]
	\node[state,initial] (s_0)   {};
	\node[state] (s_1) [right=of s_0] {};
	\node[state] (s_2) [right=of s_1] {};
	\node[state] (s_3) [right=of s_2] {};
	\path[->]
	(s_0) edge node [pos=0.5, sloped, above]{$R_2pO_2$} (s_1)
	(s_1) edge node [pos=0.5, sloped, above]{$R_2dO_1aW_2$} (s_2)
	(s_2) edge node [pos=0.5, sloped, above]{$r_2$} (s_3);
	\end{tikzpicture}
	\caption{Assumption $A$ for robot $R_1$.}
	\label{A1}
\end{figure}
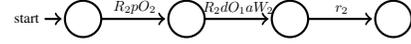

\begin{center}
	\begin{tabular}{ll}
		1 & $\langle A \rangle K^1_{MI} \langle K_{MI} \rangle $ \\
		2 & $\langle true \rangle K^2_{MI} \langle A \rangle $ \\
		\hline
		& $\langle true \rangle K^1_{MI}\vert\vert K^2_{MI}\langle K_{MI} \rangle$
	\end{tabular}
\end{center}
Here $A$ denotes an assumption about the environment (including mission plan $K^2_{MI}$ performed by robot $R_2$) in which robot $R_1$ is placed. To automatically generate appropriate assumptions, we consider the $L^*$ learning algorithm proposed in \cite{angluin1987learning}. $L^*$ creates a series of {\it observation tables} to incrementally record and maintain the information whether traces in $\Sigma^*$ belong to $U$. An observation table is a three-tuple $(S,E,T)$ consisting of: a non-empty finite set $S$ of prefix-closed traces, a non-empty finite set $E$ of suffix-closed traces and a Boolean function, called a $\emph {membership query}$, $T: (S\cup S\Sigma)E\to\{0,1\}$. Once the observation table is closed and consistent \cite{angluin1987learning}, a candidate DFA $M(S,E,T)=(Q,q_0,\delta,Q_m)$ over the alphabet $\Sigma$ is constructed. If $L(M)=U$, where $L(M)$is the {\it generated language} of $M$ \cite{cassandras2008introduction}, then the oracle returns ``True" with the current DFA $M$; otherwise, a counterexample $c\in (U-L(M))\cup(L(M)-U)$ is generated by the oracle. $L^*$ then adds all its prefixes $\overline{c}$ to $S$, which reflects the difference in next conjecture by splitting states in $M$, and $L^*$ iterates the aforementioned process to update $M$ with respect to $S$. For the purpose of compositional verification, we modify $L^*$ by using the following family of dynamical membership queries.
\begin{equation}\label{L_cv}
T_i(t) =
\begin{cases}
1,  & \mbox{if } \langle\mathcal {DFA}(t) \rangle K^1_{MI} \langle K_{MI} \rangle \mbox{ is true.} \\
0. & \mbox{otherwise}
\end{cases}
\end{equation}
where $\mathcal {DFA}(t)$ is a deterministic finite automaton that generates $\overline{t}$ and accepts $t$. In the warehouse example, an appropriate assumption $A$ for robot $R_1$ is depicted in Fig.~\ref{A1}.
\indent Next, we check whether or not $K^2_{MI}\models A$, which turns out to be true in the warehouse example. Thus one can conclude that the joint behavior of the two robots can cooperatively accomplish the global mission.
\begin{remark}\rm
	In case where $\Sigma_{MI}=\bigcup_{i\in\mathcal{N}_A} \Sigma^i_{MI}$, the compositional verification procedure essentially justifies the separability of the global mission $K_{MI}$ \cite{willner1991supervisory} with respect to $\Sigma^i_{MI}$, $i\in\mathcal{N}_{\mathcal{A}}$, i.e., $K_{MI}=||_{i\in\mathcal{N}_{\mathcal{A}}}P_i(K_{MI})$; while the assume-guarantee paradigm avoids ``state explosion" in the compositional verification. In case $K_{MI}$ is not separable, the compositional verification fails and returns a counterexample $t\in \Sigma_{MI}^*$ indicating a violation of the global mission. We present such counterexample to re-synthesize the local missions by resetting $K^i_{MI}:=K^i_{MI}-P_i(t)$. It has been shown in \cite{lindistributed} that, under the assumption that the independence relation induced by the distribution is transitive, $K_{MI}$ can always possess a non-empty separable sublanguage.
\end{remark}

\section{Bottom-up Design and Integrated Task and Motion Planning}\label{sec:CoSMoP}
\begin{figure}[!t]
	\centering
	\includegraphics[width=3.3in]{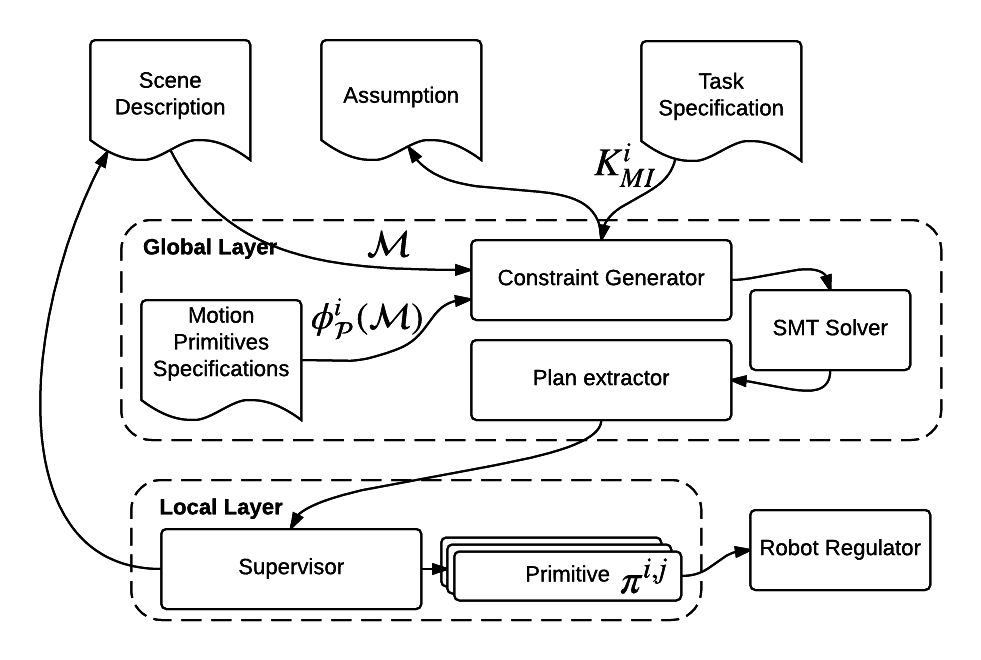}
	\caption{CoSMoP framework.}
	\label{fig:framework}
\end{figure}

This section solves the Problem 2 and illustrates it through the warehouse example. 
This section is based on extensions of our previous work \cite{CoSMoP} to multi-robot coordinations. In \cite{CoSMoP}, a bottom-up approach called CoSMoP (Composition of Safe Motion Primitives) was proposed. It features a two layer hierarchical motion planning as shown in Fig. \ref{fig:framework} for each robot. The global layer synthesizes an integrated task and motion plan for the local layer considering only geometric constraints from a given scene description $\mathcal{M}$. If this layer finds a satisfiable plan, the motion supervisor in the local layer implements a designed sequence of controller executions satisfying all kinematic and geometric constraints. 

CoSMoP solves Problem 2 in three stages. First, it designs offline a set of safe motion primitives $\mathcal{P}^{i*}$ for each robot $R_i$ to provide necessary maneuvers to complete a given task. We omit the index $i$ from now on because the controllers are identical for all robots in this paper. Second, for each primitive $\pi^j\in\mathcal{P}^*: j \in \mathcal{N}_{\mathcal{P}}$, it designs offline the corresponding specification $\phi_{\pi}^j$ in  CLTLB($\mathcal{D}$) formula to the global layer, where $\phi_{\pi}^j$ is a specification to be satisfied and the conjunction the specifications for all primitives is denominated  motion primitives specification $\phi_{\mathcal{P}}(\mathcal{M})$, i.e. $\phi_{\mathcal{P}}(\mathcal{M}) \equiv \bigwedge_{j \in \mathcal{N}_{\mathcal{P}}} \phi_{\pi}^j$. Finally, it composes a sequence of safe motion primitives to ensure the local mission $K_{MI}^i$ and the motion primitives specification $\phi_{\mathcal{P}}(\mathcal{M})$. It is solved automatically and distributively for each robot $R_i$. The following subsections will formally describe each of these steps illustrating with the Example \ref{ex:example01}.

\subsection{Design of Safe Motion Primitives}\label{sec:designofsafemp}

In the warehouse scenario, each robot $R_i$ requires five primitives such that $\mathcal{P}^* = \{\pi^1,...,\pi^5\}$, where $\pi^{1} =$ GoTo, $\pi^{2}=$ PickUp, $\pi^{3}=$ DropOff, $\pi^{4}=$ ?ObjAway (i.e. request to take an object away), $\pi^{5}=$ !ObjAway (i.e. respond that an object is taken away). 

\subsubsection{GoTo}
The controller $\pi^{1}=$ GoTo synthesizes trajectories towards a goal position based on the actual sensors readings to avoid static and moving obstacles. It can guarantee safety concerning collisions not only for the obstacles described in the scene description $\mathcal{M}$ but for other obstacles such as non-controlled agents (e.g. humans or felt down boxes) and neighbors robots. Therefore, this controller allows local and distributed trajectory synthesis that satisfies safety properties for multi-agent systems. 

The Pioneer P3-DX robot implements an embedded controller for the translational $v$ and the angular $\omega$ velocities based on the maximum acceleration $A$, deceleration $b$ and angular velocity $\Omega$. Hence, the GoTo controller is responsible for finding $v^*$ and $\omega^*$ realizable in a cycle time $T$ that specify a motion to drive the robot forward reducing the time to destination and guaranteeing the passive safe property \cite{macek2009towards}. This property means that the vehicle will never actively collide, i.e. the collision can only occur when the vehicle is stopped, and the obstacle runs into it. This property does not use the ICC (Inevitable Collision State) concept \cite{fraichard2004inevitable} because the limited range of the sensors readings and the limited knowledge assumed about the moving obstacles kinematics give limited awareness of the environment. Therefore, the controller cannot ensure that it will always find a collision-free motion. 

The robots motion $\sigma$ is a sequence of arcs $u \in \sigma$ in two-dimensional space such that the translational velocity is non-negative, the absolute value of the angular velocity is $\Omega$ and the maximum cycle time is $\epsilon$. An arc $u \in \sigma : \sigma = \{u_1, u_2, ..., u_n\}$ is specified by the translational $v$ and angular $\omega$ velocities and cycle time $T$, i.e. $u = \langle v, \omega, T \rangle$. The domain of the arcs is $\mathcal{U} = \{ u \in \mathbb{R}^3 : v \geq 0, | \omega | \leq \Omega, 0 \leq T \leq \epsilon \}$. Several types of robots can realize a motion $\sigma$, such as differential drive, Ackermann drive, single wheel drive, synchro drive, or omni drive robots \cite{braunl2008embedded}. Therefore, the trajectory realized by the Pioneer P3-DX is a motion $\sigma$. Furthermore, this motion can be modeled in d$\mathcal{L}$ to find a set $\mathcal{U}_{safe} \subseteq \mathcal{U}$ such that ensures the passive safety property. 

The primitive $GoTo$ implements an extended Dynamic Window Approach \cite{fox1997dynamic} (DWA) algorithm to avoid not only static obstacles but the ones that can be moving at a velocity up to $V$. We extend a path planning algorithm implemented in the ARNL library that synthesizes and executes trajectories to a given destination based on a map that can be generated using Mapper3\footnote{http://www.mobilerobots.com/Software/Mapper3.aspx, retrieved 05-18-2016.}. This algorithm synthesizes two trajectories: global and local trajectories. The global trajectory is a roadmap generated by an A* that considers only the static obstacles represented on the map, such as walls. The local trajectory is the trajectory implemented using a DWA algorithm that drives along the global trajectory while avoiding unmapped obstacles such as the other robots.

In summary, the DWA control searches for an arc $u^*$ at every cycle time that maximizes towards the target while avoiding a collision with obstacles that can be moving up to velocity $V$. It is organized in two steps. (i) First it searches for the dynamic window $\mathcal{U}_{dw}$ that is a range  of admissible $(v, \omega)$ pair that results in safe trajectories that the robot can realize in a short time frame $T \leq \epsilon$ such as $\mathcal{U}_{dw} \subseteq \mathcal{U}_{safe}$. A safe trajectory is the one that does not lead to a collision with an obstacle detected by the sensors readings. (ii) Then, it finds $u^* \in \mathcal{U}_{dw}$ that chooses a $(v, \omega)$ pair that maximizes the progress towards the closest next destination in the global trajectory. 

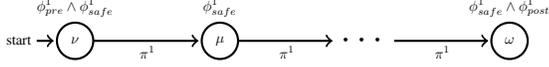
\begin{figure}[t]
	\centering
	\begin{tikzpicture}[shorten >=1pt,node distance=3.5cm,on grid,auto, bend angle=20, thick,scale=0.55, every node/.style={transform shape}]
	\node[state,initial] (s_0) [label={$\phi_{pre}^{1} \wedge \phi_{safe}^{1}$}]  {$\nu$};
	\node[state] (s_1) [right=of s_0, label={$\phi_{safe}^{1}$}] {$\mu$};
	\node (sdots) [right=of s_1] {{\Huge $\cdots$}};
	\node[state] (s_2) [right=of sdots, label={$\phi_{safe}^{1} \wedge \phi_{post}^{1}$}] {$\omega$};
	\path[->]
	(s_0) edge node [pos=0.5, sloped, below]{$\pi^{1}$} (s_1)
	(s_1) edge node [pos=0.5, sloped, below]{$\pi^{1}$} (sdots)
	(sdots) edge node [pos=0.5, sloped, below]{$\pi^{1}$} (s_2);
	\end{tikzpicture}
	\caption{Dynamic transition of the GoTo controller.}
	\label{fig:gototrans}
\end{figure}

Such control system must satisfy a safety property $\phi_{safe}^{1}$ after all its executions assuming that it starts in a state that satisfies $\phi_{pre}^{1}$ and arrives in a state that satisfies $\phi_{post}^{1}$. The Fig. \ref{fig:gototrans} shows a representation of this model in a transition system. Since $\phi_{safe}^{1}$ depends on the environment dynamics because it must be guaranteed after all executions of $\pi^{1}$, we call this property tight coupled. Furthermore, this formula is specified as a passive safety property defined in \cite{mitsch2013provably} as,
\begin{equation*}\label{eq:dwa_passive_safety_prop}
\phi_{safe}^{1} \equiv \Big(v = 0\Big) \vee \Big(\parallel p - p_o \parallel_{\infty} > \frac{v^2}{2b} + V\frac{v}{b} \Big)
\end{equation*}
where $p_r, p_o$ are the closest position of the robot and the nearest obstacle, respectively.

The added feature in the extended DWA  is that the robot will take a circular trajectory if the condition $safe$, as defined below, holds true; otherwise, it will stop. This condition is a first-order logic formula which constraints the robot state variables considering the delay caused by the cycle time. 
\begin{align*}
safe \equiv \lVert p_r - p_o \lVert_{\infty} > & \left(\frac{A}{b} + 1\right) \left(\frac{A}{2} \epsilon^2 + \epsilon v\right) \\
& +  \frac{v^2}{2b} + V \left(\epsilon + \frac{v + A \epsilon}{b}\right)
\label{eq:dwa_pf_safe}
\end{align*}

Finally, the controller is verified for $\phi_{safe}^{1}$.
\begin{thm}\label{thm:dwadl}\cite{mitsch2013provably}
	If the controller GoTo starts in a state that satisfies $\phi_{safe}^{1}$, it will always satisfies it.
	$$
	\phi_{pre}^{1} \wedge \phi_{safe}^{1} \rightarrow [(\alpha^{1})^*] \phi_{safe}^{1}
	$$ where $\phi_{pre}^{1}$ constraint only the parameters (e.g. $A > 0$, $b > 0$, $\Omega > 0$ and $\epsilon > 0$) and does not depend on any environment state, $(\alpha^{1})^*$ is the hybrid program presented in Model 1 in \cite{mitsch2013provably}, and it models the execution of the controller GoTo in d$\mathcal{L}$ for a dynamic environment with moving obstacles with maximum velocity $V$.     
\end{thm}

To guarantee passive safety, we solve the condition $safe$ and add the velocity variation with maximum acceleration $A$ for maximum cycle time $\epsilon$ (i.e. $A \epsilon$) to find the maximum value for the translational velocity setpoint $v^*$.
\begin{corollary}\label{cor:DWA_pf_maxvel}
	A circular trajectory is safe if the controller setpoint $u^* \in \mathcal{U}_{safe} : u^* = \langle v^*, \omega^*, T^* \rangle$ such as $\mathcal{U}_{safe} = \{ u \in \mathbb{R}^3 : 0 \leq v < \nu(safe), -\Omega \leq \omega \leq \Omega, 0 \leq T \leq \epsilon \}$,
	\begin{align*}
	\nu(safe) = & \begin{cases}
	v_{max} + A \epsilon & \text{if $safe$ holds true} \\
	v_{max} - b \epsilon & \text{otherwise}
	\end{cases} \\
	v_{max} = & b \cdot \sqrt{\left(\frac{A}{b} + 1\right) \epsilon^2 + \left( \frac{V}{b} \right)^2 + \frac{2 \lVert p_r - p_o \lVert_{\infty}}{b}} \\
	& - b \cdot \epsilon \left(\frac{A}{b} + 1\right)  - V
	\end{align*}
\end{corollary}
\begin{pf}
	From Model 1 in \cite{mitsch2013provably}, if a translational velocity $v$ satisfies the condition $safe$ for given position and parameters, then the acceleration can be any value between $-b$ and $A$. Since we assume that the minimum velocity for the robot is zero ($v \geq 0$), then the $safe$ condition only constraint the maximum of velocity $v$. However, the maximum translational velocity $v_{max}$ is the velocity $v$ maximum that could be reached in the next sampling time. Thus, if $safe$ holds true, the robot is allowed to accelerate up to $A$, and the maximum velocity is $v_{max} + A \epsilon$. Otherwise, the robot must brake, and the maximum velocity is $v_{max} - b \epsilon$.
\end{pf}

\subsubsection{Pick Up and Leave}
\begin{figure}[t]
	\centering
	\begin{tikzpicture}[shorten >=1pt,node distance=3.5cm,on grid,auto, bend angle=20, thick,scale=0.55, every node/.style={transform shape}]
	\node[state,initial] (s_0) [label={$\phi_{pre}^{2}$}]  {$\nu$};
	\node[state] (s_1) [right=of s_0, label={$true$}] {$\mu$};
	\node (sdots) [right=of s_1] {{\Huge $\cdots$}};
	\node[state] (s_2) [right=of sdots, label={$\phi_{pos}^{2}$}] {$\omega$};
	\path[->]
	(s_0) edge node [pos=0.5, sloped, below]{$\pi^{2}$} (s_1)
	(s_1) edge node [pos=0.5, sloped, below]{$\pi^{2}$} (sdots)
	(sdots) edge node [pos=0.5, sloped, below]{$\pi^{2}$} (s_2);
	\end{tikzpicture}
	\caption{Dynamic transition of the PickUp controller.}
	\label{fig:pickuptrans}
\end{figure}
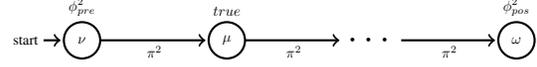

We assume that the objects in the warehouse will be picked up and dropped off by robot's gripper with a fixed robot pose, as presented in \cite{CoSMoP}. 
Hence, it must satisfy a property $\phi_{pos}^{2}$ that ensures that the robot is carrying the object after picking it up assuming that it starts in a state that satisfies $\phi_{pre}^{2}$ that guarantee that the robot is in front of the object. In contrast to the GoTo primitive, this primitive is non-tight coupled controller, meaning that $\phi_{safe}^{2}$ should be guaranteed only in the last state after finite executions $\pi^{2}$. The transition system of this controller is shown in Fig. \ref{fig:pickuptrans}. Therefore, these properties do not depend on the robot dynamics and do not need to be verified in d$\mathcal{L}$.

\subsubsection{Request and Response to Move Object Away}
We assume that the robot is stopped temporarily during the communication events. For the request controller $\pi^4$, the robot sends a request to have object $j$ moved away and waits until it receives a response message. It then continues the next planned action. The response controller $\pi^5$ means that the robot will send a message to indicate that the object $j$ is being moved. 
These primitives do not require tightly coupled safety property either, so they are not verified in d$\mathcal{L}$.

\subsection{Design of the Motion Primitives Specification}
From the local layer, we need to specify constraints for each designed controller $\pi^j \in \mathcal{P}^*: j \in \mathcal{N}_{\mathcal{P}}$ to the global layer. The conjunction of these constraints is called the motion primitive specification $\phi_{\mathcal{P}}^i(\mathcal{M})$ and is shown in the Fig. \ref{fig:framework} as one of the inputs for the constraint generator. These constraints are formulas $\phi_{\pi}^j$ in CLTLB($\mathcal{D}$) which allow the global layer to omit the kinematic constraints implemented in the controller so only geometric constraints will be considered. The CLTLB($\mathcal{D}$) is an extension of linear temporal logic (LTL) for bounded satisfiability checking (BSC) \cite{pradella2013bounded} that the models consist of temporal logic rather than transition systems; thus, the problem encoding can be more compact and elegant. Moreover, it is possible to encode CLTLB($\mathcal{D}$) into satisfiability modulo theories (SMT) \cite{bersani2010bounded} and use SMT solver to check if the specification can be satisfied. 

The formulas $\phi_{\pi}^{j}$ are specifications which constrains the states $q(k-1)^{i}$ and $q(k)^{i}$ generated in the robot $R_i$ global layer. A state $q(k)$ is assigned values for states variables in the environment at instant $k$ and the primitive taken between instants $k-1$ and $k$. Hence, this state is defined as $q(k) \in [q_{r}(k) \in Q_r] \cup [q_{b}^j(k) \in Q_b] \cup [\pi(k) \in Q_{\pi}] : k \in \mathcal{N}_{\rho}$, where $Q_r$, $Q_b$ and $Q_{\pi}$ are sequences of assigned values to robot and object states variables at each instant $k$ and assigned motion primitive to take between instants $k-1$ and $k$, respectively. Each $\phi_{\pi}^j$ must ensure that, for any plan $s^i$ for the robot $R_i$, the following two conditions hold:
\begin{itemize}
	\item For each $k\in \mathcal{N}^i$, $\phi_{safe}^{\delta,k}$ is satisfiable for at least one trajectory between  $q^i(k-1)$ and $q^i(k)$.
	\item For each $k\in \mathcal{N}^i$, $q^i(k-1) \vDash \phi_{pre}^{\delta,k}$ and $q^i(k) \vDash \phi_{post}^{\delta,k}$.
\end{itemize} The specifications $\phi_{safe}^{\delta,k}$, $\phi_{pre}^{\delta,k}$ and $\phi_{post}^{\delta,k}$ are safety properties in d$\mathcal{L}$ formulas for the primitive assigned at instant $k$ (i.e. $\delta^{i}(k) \in \mathcal{P} : k \in \mathcal{N}^i$). 
If those conditions hold true, any plan generated in the global layer that satisfies $\phi_{\mathcal{P}}(\mathcal{M})$ will guarantee the safety properties. Furthermore, the reachable states after any execution of the controller $\pi^{\delta,k} \in \mathcal{P}$ assigned in $\delta(k)$ will be constraint to satisfies initially $\phi_{pre}^{\delta,k} \wedge \phi_{safe}^{\delta,k}$, $\phi_{safe}^{\delta,k}$ after any execution of $\pi^{\delta,k}$ and it will satisfy $\phi_{post}^{\delta,k} \wedge \phi_{safe}^{\delta,k+1} \wedge \phi_{pre}^{\delta,k+1}$ before execute the next assigned controller $\pi^{\delta,k+1}$. 
\begin{thm}
	If a plan $s^i$ with size $K^i$ satisfies $\phi_{\mathcal{P}}(\mathcal{M})$ (i.e. $s^i \vDash \phi_{\mathcal{P}}(\mathcal{M})$) for a given scene description $\mathcal{M}$ and the safe motion primitives are safe (i.e. $\bigwedge_{\forall j \in \mathcal{N}_{\mathcal{P}}} \phi_{pre}^j \wedge \phi_{safe}^j \rightarrow [(\alpha^j)^*] \phi_{safe}^j$ is valid), then this plan is also safe (i.e. $s^i \vDash \bigwedge_{k \in \mathcal{N}} \phi_{safe}^{\delta,k}$).
\end{thm}
\begin{pf}
	The transition system of the plan $s^i$ is represented in the figure below.
	\begin{figure}[H]
		\centering
		\begin{tikzpicture}[shorten >=1pt,node distance=2cm,on grid,auto, bend angle=20, thick,scale=0.65, every node/.style={transform shape}]
		\node[state,initial] (s_0) [label={$\phi_{pre}^{\delta,1} \wedge \phi_{safe}^{\delta,1}$}]  {$\nu$};
		\node[state] (s_1) [right=of s_0, label={$\phi_{safe}^{\delta,1}$}] {$\mu$};
		\node (s_1dots) [right=of s_1] {{\Huge $\cdots$}};
		\node[state] (s_2) [right=of s_1dots, label={$\phi_{safe}^{\delta,1} \wedge \phi_{post}^{\delta,1} \wedge \phi_{pre}^{\delta,2} \wedge \phi_{safe}^{\delta,2}$}] {$\upsilon$};
		\node (s_2dots) [right=of s_2] {{\Huge $\cdots$}};
		\node[state] (s_3) [right=of s_2dots, label={$\phi_{safe}^{\delta,K^i} \wedge \phi_{post}^{\delta,K^i}$}] {$\omega$};
		\path[->]
		(s_0) edge node [pos=0.5, sloped, below]{$\pi^{\delta,1}$} (s_1)
		(s_1) edge node [pos=0.5, sloped, below]{$\pi^{\delta,1}$} (s_1dots)
		(s_1dots) edge node [pos=0.5, sloped, below]{$\pi^{\delta,1}$} (s_2)
		(s_2) edge node [pos=0.5, sloped, below]{$\pi^{\delta,2}$} (s_2dots)
		(s_2dots) edge node [pos=0.5, sloped, below]{$\pi^{\delta,K^i}$} (s_3);
		\end{tikzpicture}\\
		\vspace{-5 mm}
	\end{figure}    
	Since the controllers $\pi^{\delta,k}$ can be reactive, we assume that they will execute finite times until reaching the goal state that satisfies $\phi_{post}^{\delta,k}$. Thus, the safety property $\phi_{safe}^{\delta,k}$ must be ensured in the intermediate states. Let $\alpha^{\delta,k}$ is the d$\mathcal{L}$ hybrid program that models $\pi^{\delta,k}$, thus, the transition system can be modeled using d$\mathcal{L}$ formulas as in the figure below.
	\begin{figure}[H]
		\centering
		\begin{tikzpicture}[shorten >=1pt,node distance=3.5cm,on grid,auto, bend angle=20, thick,scale=0.65, every node/.style={transform shape}]
		\node[state,initial] (s_0) [label={$\phi_{pre}^{i,1} \wedge \phi_{safe}^{i,1}$}]  {$\nu$};
		\node[state] (s_1) [right=of s_0, label={$\phi_{post}^{i,1} \wedge \phi_{pre}^{i,2} \wedge \phi_{safe}^{i,2}$}] {$\mu$};
		\node (s_dots) [right=of s_1] {{\Huge $\cdots$}};
		\node[state] (s_2) [right=of s_dots, label={$\phi_{post}^{i,K}$}] {$\omega$};
		\path[->]
		(s_0) edge node [pos=0.5, sloped, below]{$[\alpha^{\delta,1*}] \phi_{safe}^{i,1}$} (s_1)
		(s_1) edge node [pos=0.5, sloped, below]{$[\alpha^{\delta,2*}] \phi_{safe}^{i,2}$} (s_dots)
		(s_dots) edge node [pos=0.5, sloped, below]{$[\alpha^{\delta,K*}] \phi_{safe}^{i,K}$} (s_2);
		\end{tikzpicture}\\
		\vspace{-5 mm}
	\end{figure}    
	The constraints defined in the specifications $\phi_{\pi}^j: j \in \mathcal{N}_{\mathcal{P}}$ can be modeled in the d$\mathcal{L}$ hybrid program using the operator $?\chi$. Therefore, the d$\mathcal{L}$ formula of resulting plan $s^i$ is,
	\begin{equation*}
	\begin{aligned}
	& \phi_{pre}^{\delta,1} \wedge \phi_{safe}^{\delta,1} \rightarrow [(\alpha^{\delta,1})^*;?(\phi_{safe}^{\delta,1});?( \phi_{post}^{\delta,1} ); \\
	& ?(\phi_{safe}^{\delta,2} \wedge \phi_{pre}^{\delta,2});(\alpha^{\delta,2})^*;?(\phi_{safe}^{\delta,2});?( \phi_{post}^{\delta,1} );\\
	& \cdots \\
	& ?(\phi_{safe}^{\delta,K^i} \wedge \phi_{pre}^{\delta,K^i});(\alpha^{\delta,K^i})^*] \phi_{safe}^{\delta,K^i} \rightarrow \phi_{post}^{\delta,K^i}
	\end{aligned}
	\end{equation*} By applying the rules $[;]$ and $[?]$ \cite{platzer2010logical}, we find the equivalent formula,
	\begin{equation}\label{eq:dlformula}
	\begin{aligned}
	& \Big(\phi_{pre}^{\delta,1} \wedge \phi_{safe}^{\delta,1} \rightarrow [(\alpha^{\delta,1})^*](\phi_{safe}^{\delta,1}) \Big) \rightarrow \phi_{post}^{\delta,1} \rightarrow \\
	& \Big(\phi_{safe}^{\delta,2} \wedge \phi_{pre}^{\delta,2} \rightarrow [(\alpha^{\delta,2})^*](\phi_{safe}^{\delta,2}) \Big) \rightarrow \phi_{post}^{\delta,2} \rightarrow \\
	& \cdots \\
	& \Big(\phi_{safe}^{\delta,K^i} \wedge \phi_{pre}^{\delta,K^i} \rightarrow [(\alpha^{\delta,K^i})^*](\phi_{safe}^{\delta,K^i}) \Big) \rightarrow \phi_{post}^{\delta,K^i} \end{aligned}
	\end{equation}    
	We know that $\bigwedge_{\forall j \in \mathcal{N}_{\mathcal{P}}} \phi_{pre}^j \wedge \phi_{safe}^j \rightarrow [(\alpha^j)^*] \phi_{safe}^j$ is valid, it means that any initial state that satisfies $\phi_{pre}^j \wedge \phi_{safe}^j$ can execute any finite times $\pi^j$ modeled as hybrid program $\alpha^j$ that it will lead to a state that is safe, i.e. satisfies $\phi_{safe}^j$. Hence, it is sufficient that the global layer find a plan $s^i$ that satisfies $\phi_{\mathcal{P}}(\mathcal{M})$ to satisfy the Eq.\ref{eq:dlformula} and, consequently, the safety property of all motion primitives for the robot $R_i$ in the scene description $\mathcal{M}$.
\end{pf}

In the next subsections, the specifications $\phi_{\pi}^j$ for each safe motion primitive are designed.

\subsubsection{GoTo}
The controller $\pi^1$ requires a tightly coupled safety property; thus we need to ensure that  $\phi_{safe}^{1}$ is satisfiable for at least one trajectory between any planned  $q^{i}(k-1)$ and $q^{i}(k)$ states. However, we assume $\phi_{pre}^{1} \equiv true$ and $\phi_{post}^{1} \equiv true$ here because these properties do not depend on the geometry or dynamics in the environment, considering that all parameters are correctly assigned (e.g. $A > 0 \equiv true$). These assumptions leave the primitive free to drive the robot to the positions required by the other primitives. Note that $\phi_{safe}^{1}$ is also an invariant property, as shown in Theorem \ref{thm:dwadl}, so we can use it to reason the existence of a safe trajectory. The global layer omits dynamic constraints; as a result, it is assumed that the minimum robot velocity is zero ($v > 0$), and the obstacles are static ($V = 0$). From the Corollary 2.1 in \cite{CoSMoP}, the Go To specification $\phi_{\pi}^1$ in CLTLB($\mathcal{D}$) should guarantee that there exists a trajectory that the robot fits in between the initial and goal state using a linear arithmetic relation. 

\begin{figure}[!t]
	\centering
	\subfloat[]{\label{fig:GoToTrajectories_BelowAbove}}
	\includegraphics[width=1.6in]{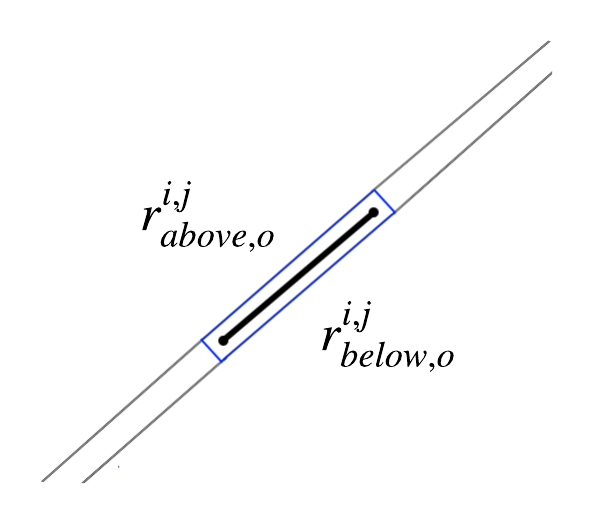}
	\subfloat[]{\label{fig:GoToTrajectories_LeftRight}}
	\includegraphics[width=1.6in]{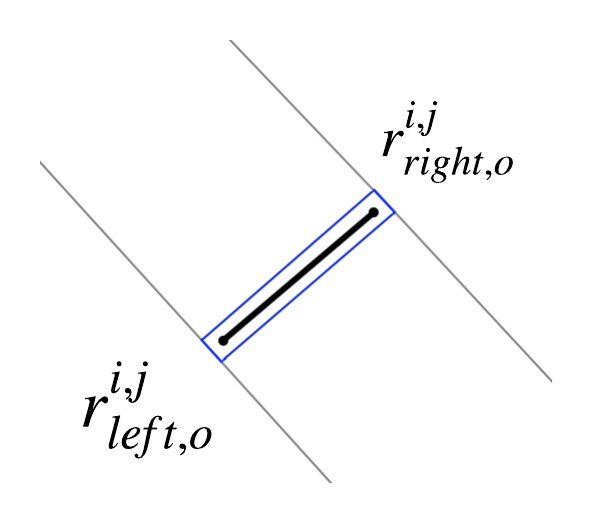}
	\caption{Regions $r_{above,o}^{i,j}$ and $r_{below,o}^{i,j}$ in (a) and $r_{left,o}^{i,j}$ and $r_{right,o}^{i,j}$ in (b).}
	\vspace{-7 mm}
	\label{fig:GoToTrajectories}
\end{figure}

\begin{thm}
	There exists a trajectory between $q^i(k-1)$ and $q^i(k)$ which satisfies the controller safety property $\phi_{safe}^{1}$, if there exists a sequence of $n$ waypoints $\{\bigcirc q_r^{i},..., \bigcirc^n q_r^{i}\}$ which the robot $R_i$ starts at the initial state $q_r^{i}$ and reaches a goal state $\bigcirc^n q_r^{i}$ that satisfies the following two conditions. First, all pairs of states $\bigcirc^{l-1} q_r^{i}$ and $\bigcirc^l q_r^{i}$ in this sequence (i.e. $l \in \{1,...,n\}$) are in region below, above, left or right of all objects line segments $o_j: j \in \mathcal{N}_{\mathcal{O}}$. Second, the environment is fair, meaning that obstacles not in the scene description do not lead the robot $R_i$ executing the primitive GoTo (i.e. $\pi^1$) to a deadlock. In this definition, it is used the \textit{a.t.t.} operator $\bigcirc^l$ as shorthands for $l$ implications of $\bigcirc$ (e.g. $\bigcirc^2 = \bigcirc \bigcirc$), where $l = 0$ means no implication.
\end{thm}
\begin{pf}
	First, since it is a fair environment, then a trajectory that is safe for static obstacles is enough to ensure the existence a safe trajectory for a dynamic environment with moving obstacles. Second, the robot states $q_r^{i}$ and $\bigcirc q_r^{i}$ constrained in one of the regions below, above, left or right of a line segment $o_j$, as shown in the Fig.~\ref{fig:GoToTrajectories}, ensure the existence of a safe trajectory. This safe trajectory may be straight line trajectory that guarantees $\phi_{safe}^{1}$ by using the Corollary 2.1 in \cite{CoSMoP}. Moreover, those regions intersect with each other, causing the existence of waypoints which is inside of more than one of these areas. Hence, it allows finding intermediate trajectories to link two states which are not in the same region. Therefore, if a sequence of waypoints $\{\bigcirc q_r^{i},..., \bigcirc^n q_r^{i}\}$ is found, there is a composition of two or more straight trajectories which leads a initial state $q_r^{i}$ to a goal state $\bigcirc^n q_r^{i}$. Furthermore, those trajectories and their composition satisfy the safety property $\phi_{safe}^{1}$ when executing the primitive GoTo. 
\end{pf}

Therefore, any pair of states $q_r^i$ and $\bigcirc q_r^i$ for robot $i$ should satisfy the following specification,
\begin{align*}
\phi_{GoTo}^{i,\mathcal{O}} \equiv \forall j \in \mathcal{N}_{\mathcal{O}} : & \Box \Big[ (\pi = \text{GoTo}) \rightarrow  \\
& r_{bellow,o}^{i,j} \vee r_{above,o}^{i,j} \vee r_{left,o}^{i,j} \vee r_{right,o}^{i,j} \Big]
\end{align*}, where
\begin{itemize}
	\item $r_{below, o}^{i,j} \equiv (isY^j)? r_{below, o}^{i,j,y} \wedge r_{below, o}^{i,j,\bigcirc y} : r_{below, o}^{i,j,x} \wedge r_{below, o}^{i,j,\bigcirc x}$;
	\item $r_{above, o}^{i,j} \equiv (isY^j)? r_{above, o}^{i,j,y} \wedge r_{above, o}^{i,j,\bigcirc y} : r_{above, o}^{i,j,x} \wedge r_{above, o}^{i,j,\bigcirc x}$;
	\item $r_{left, o}^{i,j} \equiv (\neg isY^j)? r_{left, o}^{i,j,y} \wedge r_{left, o}^{i,j,\bigcirc y} : r_{left, o}^{i,j,x} \wedge r_{left, o}^{i,j,\bigcirc x}$;
	\item $r_{right, o}^{i,j} \equiv (\neg isY^j)? r_{right, o}^{i,j,y} \wedge r_{right, o}^{i,j,\bigcirc y} : r_{right, o}^{i,j,x} \wedge r_{right, o}^{i,j,\bigcirc x}$;
	\item $r_{below, o}^{i,j,y} \equiv q_r^{i}.y \leq m_{\parallel}^j \cdot q_r^{i}.x + b_{\parallel}^j - \frac{a_i.l}{2} (1 + m_{\parallel}^j)$;
	\item $r_{below, o}^{i,j,\bigcirc y} \equiv \bigcirc q_r^{i}.y \leq m_{\parallel}^j \cdot \bigcirc q_r^{i}.x + b_{\parallel}^j - \frac{a_i.l}{2} (1 + m_{\parallel}^j)$;
	\item $r_{below, o}^{i,j,x} \equiv q_r^{i}.x \leq m_{\parallel}^j \cdot q_r^{i}.y + b_{\parallel}^j + \frac{a_i.l}{2} (1 + m_{\parallel}^j)$;
	\item $r_{below, o}^{i,j,\bigcirc x} \equiv \bigcirc q_r^{i}.x \leq m_{\parallel}^j \cdot \bigcirc q_r^{i}.y + b_{\parallel}^j + \frac{a_i.l}{2} (1 + m_{\parallel}^j)$;
	\item $r_{above, o}^{i,j,y} \equiv q_r^{i}.y \geq m_{\parallel}^j \cdot q_r^{i}.x + b_{\parallel}^j + \frac{a_i.l}{2} (1 + m_{\parallel}^j)$;
	\item $r_{above, o}^{i,j,\bigcirc y} \equiv \bigcirc q_r^{i}.y \geq m_{\parallel}^j \cdot \bigcirc q_r^{i}.x + b_{\parallel}^j + \frac{a_i.l}{2} (1 + m_{\parallel}^j)$;
	\item $r_{above, o}^{i,j,x} \equiv q_r^{i}.x \geq m_{\parallel}^j \cdot q_r^{i}.y + b_{\parallel}^j - \frac{a_i.l}{2} (1 + m_{\parallel}^j)$;
	\item $r_{above, o}^{i,j,\bigcirc x} \equiv \bigcirc q_r^{i}.x \geq m_{\parallel}^j \cdot \bigcirc q_r^{i}.y + b_{\parallel}^j - \frac{a_i.l}{2} (1 + m_{\parallel}^j)$;
	\item $r_{left, o}^{i,j,y} \equiv q_r^{i}.y \leq m_{\perp}^j \cdot q_r^{i}.x + b_{\perp,i}^j - \frac{a_i.l}{2} (1 + m_{\perp}^j)$;
	\item $r_{left, o}^{i,j,\bigcirc y} \equiv \bigcirc q_r^{i}.y \leq m_{\perp}^j \cdot \bigcirc q_r^{i}.x + b_{\perp,i}^j - \frac{a_i.l}{2} (1 + m_{\perp}^j)$;
	\item $r_{left, o}^{i,j,x} \equiv q_r^{i}.x \leq m_{\perp}^j \cdot q_r^{i}.y + b_{\perp,i}^j + \frac{a_i.l}{2} (1 + m_{\perp}^j)$;
	\item $r_{left, o}^{i,j,\bigcirc x} \equiv \bigcirc q_r^{i}.x \leq m_{\perp}^j \cdot \bigcirc q_r^{i}.y + b_{\perp,i}^j + \frac{a_i.l}{2} (1 + m_{\perp}^j)$;
	\item $r_{right, o}^{i,j,y} \equiv q_r^{i}.y \geq m_{\perp}^j \cdot q_r^{i}.x + b_{\perp,f}^j + \frac{a_i.l}{2} (1 + m_{\perp}^j)$;
	\item $r_{right, o}^{i,j,\bigcirc y} \equiv \bigcirc q_r^{i}.y \geq m_{\perp}^j \cdot \bigcirc q_r^{i}.x + b_{\perp,f}^j + \frac{a_i.l}{2} (1 + m_{\perp}^j)$;
	\item $r_{right, o}^{i,j,x} \equiv q_r^{i}.x \geq m_{\perp}^j \cdot q_r^{i}.y + b_{\perp,f}^j - \frac{a_i.l}{2} (1 + m_{\perp}^j)$;
	\item $r_{right, o}^{i,j,\bigcirc x} \equiv \bigcirc q_r^{i}.x \geq m_{\perp}^j \cdot \bigcirc q_r^{i}.y + b_{\perp,f}^j - \frac{a_i.l}{2} (1 + m_{\perp}^j)$;
	\item $isY^j \equiv |o_j.y_f - o_j.y_i| \leq |o_j.x_f - o_j.x_i|$;
	\item $m_{\parallel}^j = (isY^j)? \frac{o_j.y_f - o_j.y_i}{o_j.x_f - o_j.x_i} : \frac{o_j.x_f - o_j.x_i}{o_j.y_f - o_j.y_i}$;
	\item $m_{\perp}^j = (\neg isY^j)? -\frac{o_j.x_f - o_j.x_i}{o_j.y_f - o_j.y_i} : -\frac{o_j.y_f - o_j.y_i}{o_j.x_f - o_j.x_i}$;
	\item $b_{\parallel}^j = (isY^j)? o_j.y_i - m_{\parallel}^j \cdot o_j.x_i : o_j.x_i - m_{\parallel}^j \cdot o_j.y_i$;
	\item $b_{\perp,i}^j = (\neg isY^j)? o_j.y_i - m_{\perp}^j \cdot o_j.x_i : o_j.x_i - m_{\perp}^j \cdot o_j.y_i$;
	\item $b_{\perp,f}^j = (\neg isY^j)? o_j.y_f - m_{\perp}^j \cdot o_j.x_f : o_j.x_f - m_{\perp}^j \cdot o_j.y_f$;
	\item the operator $(relation)? value1 : value2$ returns $value1$ if $relation$ holds true, otherwise $value2$.
\end{itemize}
Note that if a constraint $r_{below, o}^{i,j}$, $r_{above,o}^{i,j}$, $r_{left,o}^{i,j}$ or $r_{right,o}^{i,j}$ holds true, then the states $q_r^i$ and $\bigcirc q_r^i$ are in the regions below, above, left or right, respectively.

And similarly we have $\phi_{GoTo}^{i,\mathcal{B}}$ to avoid colliding into objects that are not being carried (i.e. $\neg q_b^{j}.p$) and not away (i.e. $\neg g_b^{j}.a$). Thus, the initial $q_r^{i}$ and goal $\bigcirc q_r^{i}$ states should be to the left, right, below or above of all objects (i.e. $r_{left,b}^{i,j} \equiv \Big( \max (\bigcirc q_r^{i}.x, q_r^{i}.x) \leq q_b^{j}.x - d^{i,j} \Big)$, $r_{right,b}^{i,j} \equiv \Big( \min (\bigcirc q_r^{i}.x, q_r^{i}.x) \geq q_b^{j}.x + d^{i,j} \Big)$, $r_{below,b}^{i,j} \equiv \Big( \max (\bigcirc q_r^{i}.y, q_r^{i}.y) \leq q_b^{j}.y - d^{i,j} \Big)$, $r_{above,b}^{i,j} \equiv \Big( \min (\bigcirc q_r^{i}.y, q_r^{i}.y) \geq q_b^{j}.y + d^{i,j} \Big)$, where $d^{i,j} = \frac{b_j.l+a_i.l}{2}$),
\begin{align*}
\phi_{GoTo}^{i,\mathcal{B}} & \equiv \forall j \in \mathcal{N}_{\mathcal{B}} : \Box \Big[ (\pi = \text{GoTo}) \wedge \neg q_b^{j}.p \wedge \neg q_b^{j}.a \rightarrow  \\
& r_{left,b}^{i,j} \vee r_{right,b}^{i,j} \vee r_{bellow,b}^{i,j} \vee r_{above,b}^{i,j} \Big]
\end{align*}
Finally, the robot should't change any object state (i.e. $p_{static}^{l} \equiv \bigcirc q_b^{l}.p = q_b^{l}.p$ and $a_{static}^{l} \equiv \bigcirc q_b^{l}.a = q_b^{l}.a$) when executing $Go To$, so, we have,
\begin{align*}
\phi_{GoTo}^{i} \equiv & \Box \Big[ \pi = \text{GoTo} \rightarrow \bigwedge_{l \in \mathcal{N}_{\mathcal{B}}} \Big(p_{static}^{l} \wedge a_{static}^{l}\Big)\Big] \wedge \\
& \phantom{\Box \Big[ \pi = GoTo \rightarrow} \phi_{GoTo}^{i,\mathcal{O}} \wedge \phi_{GoTo}^{i,\mathcal{B}} 
\end{align*}

\subsubsection{PickUp and DropOff}
We assume that the robot can only pick up the object when $q_r^2.\alpha = 0\degree$. Hence, to pick an object up, the robot cannot be carrying any object (i.e. $\neg q_b^{l}.p$) and will carry the object $j$ (i.e. $\bigcirc p_{carry}^{j,l} \equiv (j=l \rightarrow \bigcirc q_b^{l}.p) \wedge (j \neq l \rightarrow \neg \bigcirc q_b^{l}.p)$). Also, the robot initial and goal states will not change (i.e. $r_{static}^i \equiv (q_r^{i}.x = \bigcirc q_r^{i}.x) \wedge (q_r^{i}.y = \bigcirc q_r^{i}.y) \wedge (q_r^{i}.{\alpha} = \bigcirc q_r^{i}.{\alpha})$) and it will be posing in front of object (i.e. $r_{object}^{i,j} \equiv (q_r^{i}.\alpha = 0.0) \wedge (q_r^{i}.y = q_b^{j}.y) \wedge (q_r^{i}.x = q_b^{j}.x - d)$),
\begin{align*}
& \phi_{PickUp}^{i} \equiv \forall j \in \mathcal{N}_{\mathcal{B}}: \Box \Big[ \Big(\pi = PickUp_j\Big) \rightarrow \\ 
& \bigwedge_{\forall l \in \mathcal{N}_{\mathcal{B}}} (\neg q_b^{l}.p \wedge \bigcirc p_{carry}^{j,l}) \wedge r_{static}^i \wedge r_{object}^{i,j}\Big]
\end{align*}
Accordingly, we drop the object off at the same angle. Thus, the robot should be carrying the object $j$ (i.e. $p_{carry}^{j,l} \equiv (j=l \rightarrow q_b^{l}.p) \wedge (j \neq l \rightarrow \neg q_b^{l}.p)$) and, then, not (i.e. $\neg \bigcirc q_b^{l}.p$). Moreover, the robot will not change its the initial and final states (i.e. $r_{static}^i$) and the object will be left next to it at $0^o$ (i.e. $b_{left}^{i,j} \equiv (q_r^{i}.\alpha = 0.0) \wedge (\bigcirc q_b^{j}.y = q_r^{i}.y) \wedge (\bigcirc q_b^{j}.x = q_r^{i}.x + d)$).
However, we cannot leave the object over other objects. Therefore, the next object position should be to the left, right, below or above of all other objects (i.e. $b_{left,b}^{j,l} \equiv \Big( \bigcirc q_b^{j}.y \leq \bigcirc q_b^{l}.y - d_b^{j,l} \Big)$, $b_{right,b}^{j,l} \equiv \Big( \bigcirc q_b^{j}.y \geq \bigcirc q_b^{l}.y + d_b^{j,l} \Big)$, $b_{below,b}^{j,l} \equiv \Big( \bigcirc q_b^{j}.x \leq \bigcirc q_b^{l}.x - d_b^{j,l} \Big)$, $b_{above,b}^{j,l} \equiv \Big( \bigcirc q_b^{j}.x \geq \bigcirc q_b^{l}.x + d_b^{j,l} \Big)$, where $d_b^{j,l} = \frac{b_j.l+b_l.l}{2}$),
\begin{align*}
& \phi_{DropOff}^{i, \mathcal{B}} \equiv \forall j,l \in \mathcal{N}_{\mathcal{B}}, j \neq l :  \\
& \Box \Big[ \Big((\pi = \text{DropOff}_j) \wedge (\neg q_b^{l}.p) \wedge (\neg q_b^{l}.a) \Big)\rightarrow \\
& \Big(b_{left,b}^{j,l} \vee b_{right,b}^{j,l} \vee b_{below,b}^{j,l} \vee b_{above,b}^{j,l} \Big)  \Big]
\end{align*}
Similarly, neither over an obstacle. Hence the object should be left to the left, right, below or above of all obstacles ,
\begin{align*}
\phi_{DropOff}^{i,\mathcal{O}} \equiv & \forall j \in \mathcal{B}, l \in \mathcal{O} : \Box \Big[ (\pi = \text{DropOff}_j) \rightarrow \\
& b_{left,o}^{j,l} \vee b_{right,o}^{j,l} \vee b_{below,o}^{j,l} \vee b_{above,o}^{j,l} \Big]
\end{align*}, where
\begin{itemize}
	\item $b_{below, o}^{j,l} \equiv (isY^j)? b_{below, o}^{j,l,\bigcirc y} : b_{below, o}^{j,l,\bigcirc x}$;
	\item $b_{above, o}^{j,l} \equiv (isY^j)? b_{above, o}^{j,l,\bigcirc y} : b_{above, o}^{j,l,\bigcirc x}$;
	\item $b_{left, o}^{j,l} \equiv (\neg isY^j)? b_{left, o}^{j,l,\bigcirc y} : b_{left, o}^{j,l,\bigcirc x}$;
	\item $b_{right, o}^{j,l} \equiv (\neg isY^j)? b_{right, o}^{j,l,\bigcirc y} : b_{right, o}^{j,l,\bigcirc x}$;
	\item $b_{below, o}^{j,l,\bigcirc y} \equiv \bigcirc q_b^{j}.y \leq m_{\parallel}^l \cdot \bigcirc q_b^{j}.x + b_{\parallel}^l - \frac{b_j.l}{2} (1 + m_{\parallel}^l)$;
	\item $b_{below, o}^{j,l,\bigcirc x} \equiv \bigcirc q_b^{j}.x \leq m_{\parallel}^l \cdot \bigcirc q_b^{j}.y + b_{\parallel}^l + \frac{b_j.l}{2} (1 + m_{\parallel}^l)$;
	\item $b_{above, o}^{j,l,\bigcirc y} \equiv \bigcirc q_b^{j}.y \geq m_{\parallel}^l \cdot \bigcirc q_b^{j}.x + b_{\parallel}^l + \frac{b_j.l}{2} (1 + m_{\parallel}^l)$;
	\item $b_{above, o}^{j,l,\bigcirc x} \equiv \bigcirc q_b^{j}.x \geq m_{\parallel}^l \cdot \bigcirc q_b^{j}.y + b_{\parallel}^l - \frac{b_j.l}{2} (1 + m_{\parallel}^l)$;
	\item $b_{left, o}^{j,l,\bigcirc y} \equiv \bigcirc q_b^{j}.y \leq m_{\perp}^l \cdot \bigcirc q_b^{j}.x + b_{\perp,i}^l - \frac{b_j.l}{2} (1 + m_{\perp}^l)$;
	\item $b_{left, o}^{j,l,\bigcirc x} \equiv \bigcirc q_b^{j}.x \leq m_{\perp}^l \cdot \bigcirc q_b^{j}.y + b_{\perp,i}^l + \frac{b_j.l}{2} (1 + m_{\perp}^l)$;
	\item $b_{right, o}^{j,l,\bigcirc y} \equiv \bigcirc q_b^{j}.y \geq m_{\perp}^l \cdot \bigcirc q_b^{j}.x + b_{\perp,f}^l + \frac{b_j.l}{2} (1 + m_{\perp}^l)$;
	\item $b_{right, o}^{j,l,\bigcirc x} \equiv \bigcirc q_b^{j}.x \geq m_{\perp}^l \cdot \bigcirc q_b^{j}.y + b_{\perp,f}^l - \frac{b_j.l}{2} (1 + m_{\perp}^l)$;
\end{itemize}
Therefore,
\begin{align*}
& \phi_{DropOff}^{i} \equiv \forall j \in \mathcal{N}_{\mathcal{B}}: \Box \Big[ \Big(\pi = \text{DropOff}_j\Big) \rightarrow \\
& \bigwedge_{\forall l \in \mathcal{N}_{\mathcal{B}}} \Big((p_{carry}^{j,l}) \wedge (\neg \bigcirc q_b^{l}.p) \wedge (a_{static}^{l})\Big) \wedge r_{static}^i \wedge b_{left}^{i,j}\Big] \\
& \wedge \phi_{DropOff}^{i, \mathcal{B}} \wedge \phi_{DropOff}^{i,\mathcal{O}}
\end{align*}

Finally, we allow changing the object position only if the robot leaves it.
\begin{align*}
\phi_{carry}^{i} \equiv \forall j \in \mathcal{N}_{\mathcal{B}} : & \Box \Big[ \Big( (\pi \neq \text{DropOff}_j) \rightarrow \\
& (\bigcirc q_b^{j}.x = q_b^{j}.x) \wedge (\bigcirc q_b^{j}.y = q_b^{j}.y) \Big) \Big]
\end{align*}

\subsubsection{Request and Response to Move Object Away}
The abstraction of request controller $u^4$ constraints that the object state $q_b^j.a$ must change from $false$ to $true$ (i.e. $a_{change}^{j,l} \equiv (\neg q_b^{j}.a) \wedge (j=l \rightarrow \bigcirc q_b^{j}.a) \wedge (j\neq l \rightarrow \neg \bigcirc q_b^{j}.a)$) before continuing the rest of the task. Also the robot is static (i.e. $r_{static}^i$ and $p_{static}^{l}$) in a position that provide enough space for other robots to pick the requested object up (i.e. robot $R_i$ away from object $i$, $r_{away}^{i,j} \equiv \Big(q_r^{i}.x \leq q_b^{j} - (b_j.l + a_i.l)\Big) \vee \Big(q_r^{i}.x \geq q_b^{j} + \frac{b_j.l}{2}\Big) \vee \Big(q_r^{i}.y \leq q_b^{j} - \frac{a_i.l}{2}\Big) \vee \Big(q_r^{i}.y \geq q_b^{j} + \frac{a_i.l}{2}\Big)$). Thus,
\begin{align*}
& \phi_{Req}^{i} \equiv \forall j \in \mathcal{N}_{\mathcal{B}} : \Box \Big[ \Big[ (\pi = \text{Req}_j) \rightarrow \\ 
& \bigwedge_{l \in \mathcal{N}_{\mathcal{B}}} \Big(p_{static}^{l} \wedge a_{change}^{j,l}\Big) \wedge r_{static}^i \wedge r_{away}^{i,j} \Big]
\end{align*}
For the response controller $u^5$, its abstraction also constraints that the robot is static (i.e. $r_{static}^i$ and $p_{static}^{l}$ and $a_{satic}^{l} \equiv (\bigcirc q_b^{l}.a = q_b^{l}.a)$). Furthermore, the object must already be picked up (i.e. $q_b^{j}.p$) and the robot state must eventually provide enough space for another robot to pick up their objects (i.e. $r_{away}^{i,j}$). 
Hence,
\begin{align*}
& \phi_{Res}^{i} \equiv \forall j \in \mathcal{N}_{\mathcal{B}} : \Box \Big[(\pi = \text{Res}_j) \rightarrow \\
& q_b^{j}.p \wedge \bigwedge_{l \in \mathcal{N}_{\mathcal{B}}} \Big(p_{static}^{l} \wedge a_{satic}^{l} \Big) \wedge r_{static}^i \wedge \diamondsuit r_{away}^{i,j} \Big]
\end{align*}

\subsection{Composition of safe motion primitives}    
The composition of safe motion primitives is implemented in the global layer as shown in the Fig. \ref{fig:framework} based on the generated local mission plan $K^i_{MI}$.  Specifically, we assume that the following are given,
\begin{itemize}
	\item a local mission $K^i_{MI}$ presented by a DFA;
	\item a scene description $\mathcal{M}$;
	\item a motion primitive specification $\phi_{\mathcal{P}}^{i}(\mathcal{M}) \equiv \bigwedge_{j \in \mathcal{N}_{\mathcal{U}^{i}}} \phi_{\Pi^{j}}$.    
\end{itemize}
First, we encode $K^i_{MI}$ as a CLTLB($\mathcal{D}$) specification to the SMT solver online. With these encoding, we can check if  $K^i_{MI}$ is satisfiable in the scene description $\mathcal{M}$ for available safe motion primitives. If yes, we find a roadmap $\langle Q_r^{i}, \delta^{i} \rangle$ with minimum trace length $K^i$ at the global layer that can be executed at the local layer. If the bottom-up motion planning finds that the $K_{MI}^i$ is not feasible, it will provide feedback to require and initiate inter-agent coordination in the top-down mission planning, resulting in the re-allocation of the local missions $K_{MI}^i$. 

\subsubsection{Encoding of the local mission plan $K^i_{MI}$}
We encode each event $\sigma\in \Sigma_{MI}^{i}$ into a symbol that represents a CLTLB($\mathcal{D}$) formula that describes which reactive motion controllers can be executed. For the Example \ref{ex:example01}, the events $\{R_ipOj$, $R_idO_jaW_k$, $r_i$, $O_jAway\}$ can be encoded as:
\begin{flalign*}
& R_ipO_j \equiv  \Big( (\pi = \text{GoTo}) \vee \bigvee_{\forall l \in \mathcal{N}_{\mathcal{B}}, l \neq j}(\pi = \text{Req}_l) \Big)\mathbf{U}\\ & \Big(\pi = \text{DropOff}_j\Big)\\
& R_idO_jaW_1 \equiv
\Big( (\pi = \text{GoTo}) \vee \bigvee_{\forall l \in \mathcal{N}_{\mathcal{B}}, l \neq j} (\pi = \text{Req}_l) \Big)\mathbf{U}\\ & \Big( \neg q_b^{j}.p \wedge (-1500 \leq q_b^{j}.x \leq -1000) \wedge (2000 \leq q_b^{j}.y \leq 2500) \Big) \\
& R_idO_jaW_2 \equiv  \Big( (\pi = \text{GoTo}) \vee \bigvee_{\forall l \in \mathcal{N}_{\mathcal{B}}, l \neq j} (\pi = Req_l) \Big) \mathbf{U}\\& \Big( \neg q_b^{j}.p \wedge
(1500 \leq q_b^{j}.x \leq 1000) \wedge (2000 \leq q_b^{j}.y \leq 2500) \Big) \\
\end{flalign*}
\begin{flalign*}
& r_i \equiv  \Big(\Big( (\pi = \text{GoTo}) \vee \bigvee_{l \in \mathcal{N}_{\mathcal{B}}}(\pi = \text{Req}_l)\Big) \mathbf{U}\\
& \Big( (q_r^{i}.{\alpha} = a_i.q_{r,0}.{\alpha}) \wedge
q_r^{i}.x = a_i.q_{r,0}.x) \wedge (q_r^{i}.y = a_i.q_{r,0}.y) \Big) \\
&!O_jAway \equiv
\Big( \pi = \text{GoTo} \Big) \mathbf{U} \Big( ( \pi = \text{DropOff}_j ) \mathbf{U} ( \pi = \text{Res}_j )\Big)
\end{flalign*}

Then we encode the sequential DFA mission plan with  nested until operator $\mathbf{U}$. For example,  $K^1_{MI}$ in Fig. \ref{K1} is encoded in CLTLB($\mathcal{D}$) as $(R_1pO_1) \mathbf{U} \Big((R_1dO_1aW_1) \mathbf{U} (r_1)\Big)$.

%
%
%
%

\subsubsection{Encoding to SMT solver}\label{sec:comp}
The motion primitive specifications $\phi_{\mathcal{P}}^{i}(\mathcal{M})$ for each robot $R_i$ are the conjunctions of the specifications from each single motion primitive. For the Example \ref{ex:example01}, the specification is,
\begin{align*}
\phi_{\mathcal{P}}^{i}(\mathcal{M}) \equiv & \phi_{GoTo}^{i} \wedge \phi_{PickUp}^{i} \wedge \phi_{Leave}^{i}  \wedge \phi_{carry}^{i} \wedge \\
& \phi_{Req}^{i} \wedge \phi_{Res}^{i}
\end{align*}
Now we can compose the motion primitives by encoding the local mission plan $K^i_{MI}$ and the motion primitive specifications $\phi_{\mathcal{P}}^{i}(\mathcal{M})$ to Z3 SMT solver \cite{de2008z3}. If the specifications are satisfiable, the SMT solver will output a feasible plan $s^i$. 

Each state variable defined in the CLTL($\mathcal{D}$) specifications are encoded as an array of variables in the SMT solver, because Z3 is a decision procedure for the combination of quantifier-free first-order logic with theories for linear arithmetic \cite{de2008z3}. For example, a robot $R_i$ state $q_r^i.x$ is encoded as an array $q_r.x[k]$ such that $k \in \mathcal{N}_{\rho}$. A object state is encoded as a two dimensional array such that each element $q_b^j.x$ is $q_b[j].x[k]$, where $j \in \mathcal{N}_{\mathcal{B}}$ and $k \in \mathcal{N}_{\rho}$. Further, each motion primitives $\pi(k) \in Q_{\pi}$ will be an array such that each element is $\pi[k]$, where $k \in \mathcal{N}$ because we do not assign any value at initial state. 

The \textit{a.t.t.} operator $\bigcirc$ can be encoded by adding or subtracting the array index, for instance, $\bigcirc q_r.x \equiv q_r.x[k+1]$ at instant $k$. Therefore, a state formula $\psi$, which is a formula defined as $\psi \equiv p \mid R(\varphi_1, \varphi_2,..., \varphi_n) \mid \neg \psi \mid \psi_1 \wedge \psi_2$, can be encoded to quantifier-free first-order logic formulas $\Psi[k]$, where $k \in \mathcal{N}_{\rho}$ is the instant that $\psi$ holds true. For instance, if $\psi \equiv q_b^0.p$, then $\Psi[2]$ holds true if $q_b^0.p$ holds true at instant $2$.  

Encoding temporal logic quantifiers to first order logic requires quantifiers $\forall$ and $\exists$ in relation to the time instants. The quantifier $\forall k \in \mathcal{N}_{\rho} : \Psi[k]$ can be implemented using for loop. The $\exists k \in \mathcal{N}_{\rho} : \Psi[k]$ can be encoded by using an auxiliary variable $j$ such as $\forall k \in \mathcal{N}_{\rho}: (k = j) \rightarrow \Psi[k] \wedge j \in \mathcal{N}_{\rho}$ and, then, also encoded using a for loop. Therefore, we can encode CLTLB($\mathcal{D}$) quantifiers to Z3, for example,

\begin{itemize}
	\item $\bigcirc^j \psi \Longleftrightarrow j \in \mathcal{N}_{\rho} \wedge \Psi[j]$
	\item $\psi_1 \mathbf{U} \psi_2 \Longleftrightarrow \begin{cases}
	\Big(\bigwedge_{k \in \mathcal{N}_{\rho}} \Big[(k < j \rightarrow \Psi_1[k]) \wedge \\
	(k = j \rightarrow \Psi_2[k])\Big] \wedge j \in \mathcal{N}_{\rho}
	\end{cases}$
	\item $\Box \psi \Longleftrightarrow \bigwedge_{k \in \mathcal{N}_{\rho}} \Psi[k]$
	\item $\diamondsuit \psi \Longleftrightarrow \bigwedge_{k \in \mathcal{N}_{\rho}} \Big[k = j \rightarrow \Psi[k]\Big] \wedge j \in \mathcal{N}_{\rho}$
	\item $Last [\psi] \Longleftrightarrow \Psi[K]$
	\item $\psi_1 \mathbf{U} (\psi_2 ... \mathbf{U} \psi_N) \Longleftrightarrow \begin{cases}
	\bigwedge_{k \in \mathcal{N}_{\rho}} \Big[(k < j_1 \rightarrow \Psi_1[k]) \wedge \\
	(j_1 \leq k < j_2 \rightarrow \Psi_2[k]) \wedge \\
	\cdots \wedge (k = j_N \rightarrow \Psi_N[k])\Big] \wedge \\
	j_1,...,j_N \in \mathcal{N}_{\rho} \wedge \\
	j_1 < j_2 < \cdots < j_N
	\end{cases}$
\end{itemize}

Finally, let $\varphi_1$ and $\varphi_2$ be \textit{a.t.t.}'s, the functions $\max(\varphi_1,\varphi_2)$ and $\min(\varphi_1,\varphi_2)$ are encoded with SMT function $ite$, i.e. $\max(\varphi_1,\varphi_2) \equiv ite(\varphi_1 > \varphi_2, \varphi_1,\varphi_2)$ and $\min(x,y) \equiv ite(\varphi_1 < 
\varphi_2,\varphi_1,\varphi_2)$. Now, we can define a task specification in CLTLB($\mathcal{D}$) and find an integrated task and motion plan $s^i$ for the scenario in the Example \ref{ex:example01} as shown below.

\begin{example}\rm
	If we encode the mission plans in Fig. \ref{fig:sup} to the scene description in Example \ref{ex:example01}, the local motion plan for robots $1$ and $2$ will be,
	{\small 
		\begin{align*}
		s^1 = \{&\langle \pi_4, (-2000, -2000, 0) \rangle, \langle \pi_1, (1750, -1000, 0) \rangle,\\
		& \langle \pi_2, (1750, -1000, 0) \rangle, \langle \pi_1, (-1250, -200, 0) \rangle, \\ 
		& \langle \pi_1, (-1501, 2000, 0) \rangle, \langle \pi_3, (-1501, 2000, 0) \rangle, \\ 
		& \langle \pi_1, (-2000, -1000, 0) \rangle\} \\
		s^2 = \{&\langle \pi_1, (1650, -1000, 0) \rangle, \langle \pi_2, (1650, -1000, 0) \rangle,\\
		& \langle \pi_1, (1201, 2000, 0) \rangle, \langle \pi_3, (1201, 2000, 0) \rangle, \\
		& \langle \pi_1, (-2000, -2000, 0) \rangle\}
		\end{align*}
	}
	However, the robot $R_1$ plan $s^1$ requires another robot to move the object $2$ away. Therefore, the request event $?O_2Away$ must be added to the mission plan $K_{MI}^1$. This feedback information will be used in the mission planning level to check the feasibility of the re-allocated mission plans. To establish the inter-robot coordination, we first add a pair of request-response events $(?O_2Away, !O_2Away)$ to the local missions in order to maintain well-posedness of the multi-robot system. The modified specifications, deemed as $\tilde K^1_{MI}$ and $\tilde K^2_{MI}$,  are illustrated in Fig. \ref{fig:supervisor2}, respectively. Next, we recall the compositional verification procedure stated in Section III to examine whether or not $\tilde M_1\vert\vert \tilde M_2 \models K_{MI}$, where $\tilde M_i=\tilde K^i_{MI}$, $i=1,2$. It turns out that the new missions are satisfiable and the global mission can be accomplished jointly.
	\begin{figure}[H]
		\centering
		\subfloat[$\tilde K^1_{MI}$]{\label{S1}    
			\begin{tikzpicture}[shorten >=1pt,node distance=2.5cm,on grid,auto, bend angle=20, thick,scale=0.55, every node/.style={transform shape}]
			\node[state,initial] (s_0)   {};
			\node[state] (s_1) [right=of s_0] {};
			\node[state] (s_2) [right=of s_1] {};
			\node[state] (s_3) [right=of s_2] {};
			\node[state] (s_4) [right=of s_3] {};
			\path[->]
			(s_0) edge node [pos=0.5, sloped, above]{$?O_2Away$} (s_1)
			(s_1) edge node [pos=0.5, sloped, above]{$R_1pO_1$} (s_2)
			(s_2) edge node [pos=0.5, sloped, above]{$R_1dO_1aW_1$} (s_3)
			(s_3) edge node [pos=0.5, sloped, above]{$r_1$} (s_4);
			\end{tikzpicture}}\\
		\subfloat[$\tilde K^2_{MI}$]{\label{S2}
			\begin{tikzpicture}[shorten >=1pt,node distance=2.5cm,on grid,auto, bend angle=20, thick,scale=0.55, every node/.style={transform shape}]
			\node[state,initial] (s_0)   {};
			\node[state] (s_1) [right=of s_0] {};
			\node[state] (s_2) [right=of s_1] {};
			\node[state] (s_3) [right=of s_2] {};
			\node[state] (s_4) [right=of s_3] {};
			\path[->]
			(s_0) edge node [pos=0.5, sloped, above]{$R_2pO_2$} (s_1)
			(s_1) edge node [pos=0.5, sloped, above]{$!O_2Away$} (s_2)
			(s_2) edge node [pos=0.5, sloped, above]{$R_2dO_2aW_2$} (s_3)
			(s_3) edge node [pos=0.5, sloped, above]{$r_2$} (s_4);        
			\end{tikzpicture}}
		\caption{New local missions for each robot.}
		\label{fig:supervisor2}
		\vspace{-5 mm}
	\end{figure}
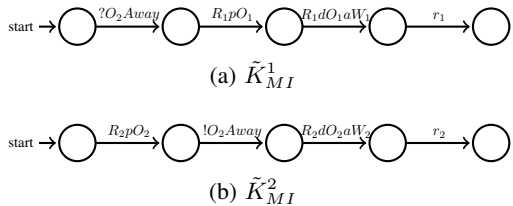
	Hence, a new plan $s^2$ is generated for $R_2$ that satisfies $\tilde K^2_{MI}$ including a response primitive $\pi^5$,
	{\small        
		\begin{align*}
		s^2 = \{&\langle \pi_1, (1650, -1000, 0) \rangle, \langle \pi_2, (1650, -1000, 0) \rangle,\\
		& \langle \pi_5, (1650, -1000, 0) \rangle, \langle \pi_1, (1201, 2000, 0) \rangle, \\ 
		& \langle \pi_3, (1201, 2000, 0) \rangle, \langle \pi_1, (-2000, -2000, 0) \rangle\}
		\end{align*}
	}    
\end{example}

\begin{figure}[h]
	\centering
	\includegraphics[width=1.5in]{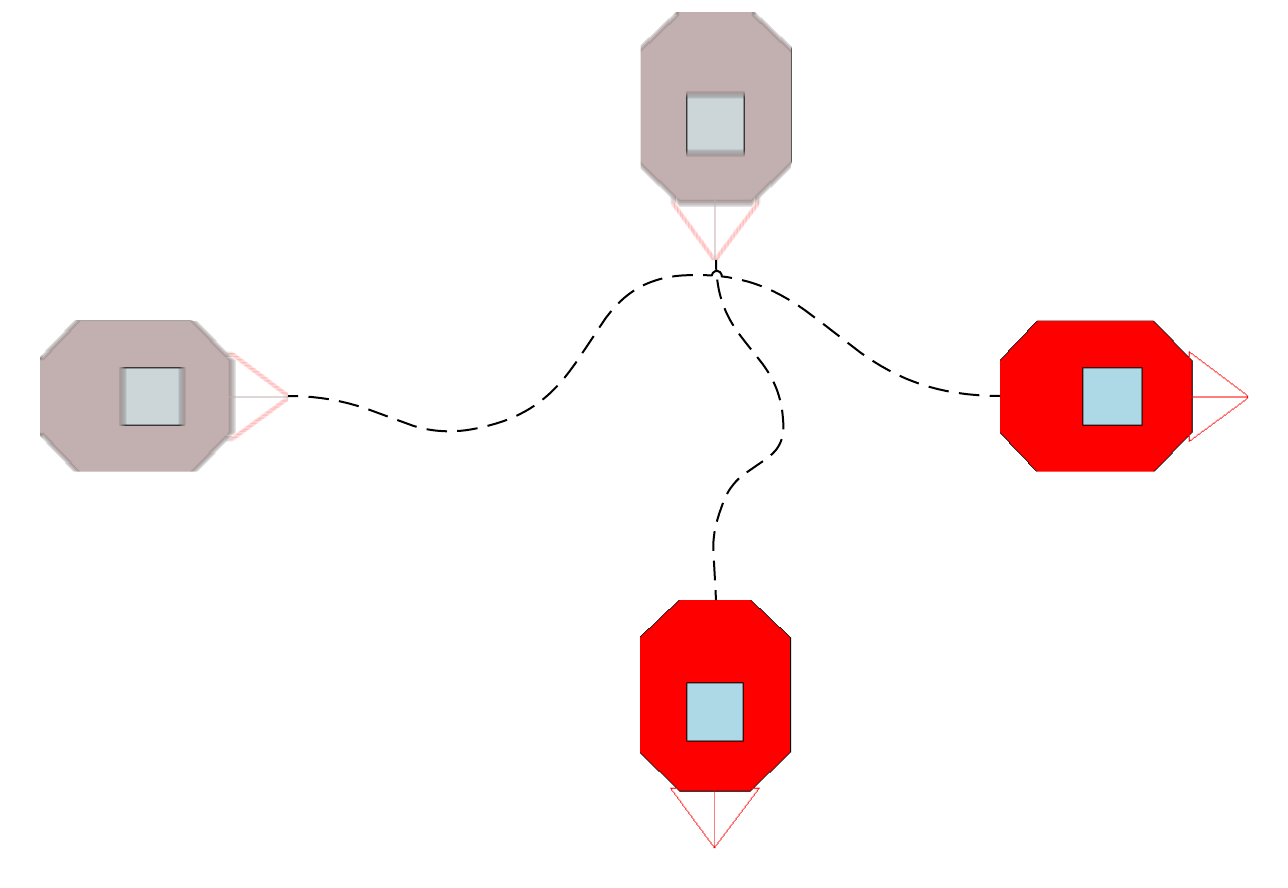}
	\caption{An illustration of trajectories generated by GoTo when two robots cross each other. Gray robots are initial and red are last positions. Circular trajectories are assigned towards the goal position while avoiding the collision, where the translational velocity is adjusted to ensure the safety property.}
	\label{fig:dwa_trajectory}
	
\end{figure}

Note that those plans are safe to moving obstacles including other agents. For example, when the robot $R_1$ is executing the primitive GoTo to go to pose $(1750, -1000, 0)$ (i.e. $\langle \pi_1, (1750, -1000, 0) \rangle$), it may encounter the robot $R_2$ executing GoTo to go to pose $(1201, 2000, 0)$ (i.e. $\langle \pi_1, (1201, 2000, 0) \rangle$). Thus, those robots will generate locally safe circular trajectories to avoid each other with low computation as shown in Fig.~\ref{fig:dwa_trajectory}. Additionally, if the environment is fair, those trajectories will lead them to the goal position. If it is not and our assumption cannot be guaranteed, the robots will always be in the safe state. Hence, we can update the scene description  $\mathcal{M}$ and search for new plans $s^i$ at current state in a receding horizon strategy. 

\begin{example}\rm
	Now, we present a scenario that tests the scalability of the proposed approach. The scenario includes a square room with 10 robots and 10 objects as defined above,
	\begin{flalign*}
	\mathcal{O} = & \Big[\langle (-5000,-5000), (5000,-5000) \rangle, \\
	& \langle (5000,-5000), (5000,5000) \rangle, \\
	& \langle (5000,5000), (-5000,5000) \rangle, \\
	& \langle (-5000,5000), (-5000,-5000) \rangle \Big] \\
	\end{flalign*}
	\begin{flalign*}
	\mathcal{A} = & \Big[ \langle 500, (-4000, 0, 0.0) \rangle, \langle 500, (-4000, -1000, 0.0) \rangle, \\
	& \langle 500, (-4000, -2000, 0.0) \rangle,  \langle 500, (-4000, -3000, 0.0) \rangle, \\
	& \langle 500, (-4000, -4000, 0.0) \rangle,  \langle 500, (-3000, 0, 0.0) \rangle, \\
	& \langle 500, (-3000, -1000, 0.0) \rangle,  \langle 500, (-3000, -2000, 0.0) \rangle, \\
	& \langle 500, (-3000, -3000, 0.0) \rangle,  \langle 500, (-3000, -4000, 0.0) \rangle\Big] \\
	\end{flalign*}
	\begin{flalign*}
	\mathcal{B} = & \Big[\langle 100, (4000, -1000, false, false) \rangle, \\
	& \langle 100, (3800, -1000, false, false) \rangle, \\
	& \langle 100, (3600, -1000, false, false) \rangle, \\
	& \langle 100, (4000, -2000, false, false) \rangle, \\
	& \langle 100, (3800, -2000, false, false) \rangle, \\
	& \langle 100, (3600, -2000, false, false) \rangle, \\        
	& \langle 100, (4000, -3000, false, false) \rangle, \\
	& \langle 100, (3800, -3000, false, false) \rangle, \\
	& \langle 100, (4000, -4000, false, false) \rangle, \\
	& \langle 100, (3800, -4000, false, false) \rangle\Big].
	\end{flalign*}
	The main challenges added in this scenario are: first, we added strings of three objects that should require coordination with more then two robots to pick up them; second, the room is small enough to lead the robots to often cross the way of each other. Hence, we not only add number of robots, which also adds computational effort, but we add complexities in the problem. 
	
	The global mission is decomposed in a set of local missions $K_{MI}^i : i \in \mathcal{N}_{\mathcal{A}} = \{1,...,10\}$ such as,
	\begin{figure}[H]
		\centering
		\begin{tikzpicture}[shorten >=1pt,node distance=2.5cm,on grid,auto, bend angle=20, thick,scale=0.55, every node/.style={transform shape}]
		\node[state,initial] (s_0)   {};
		\node[state] (s_1) [right=of s_0] {};
		\node[state] (s_2) [right=of s_1] {};
		\node[state] (s_3) [right=of s_2] {};
		\path[->]
		(s_0) edge node [pos=0.5, sloped, above]{$R_ipO_i$} (s_1)
		(s_1) edge node [pos=0.5, sloped, above]{$R_idO_iaW_i$} (s_2)
		(s_2) edge node [pos=0.5, sloped, above]{$r_i$} (s_3);
		\end{tikzpicture}
		\vspace{-5 mm}
	\end{figure}
	where,
	\begin{flalign*}
	& dO^{(i,j)} \equiv \Big( (\pi = \text{GoTo}) \vee \bigvee_{\forall l \in \mathcal{N}_{\mathcal{B}}, l \neq j} (\pi = \text{Req}_l) \Big) \\
	& R_idO_jaW_1 \equiv dO^{(i,j)} \mathbf{U} (\neg q_b^{j}.p \wedge q_b^{j}.(x, y) = (-4250, 4000) \\
	& R_idO_jaW_2 \equiv dO^{(i,j)} \mathbf{U} (\neg q_b^{j}.p \wedge q_b^{j}.(x, y) = (-3250, 4000) \\
	& R_idO_jaW_3 \equiv dO^{(i,j)} \mathbf{U} (\neg q_b^{j}.p \wedge q_b^{j}.(x, y) = (-2250, 4000) \\
	& R_idO_jaW_4 \equiv dO^{(i,j)} \mathbf{U} (\neg q_b^{j}.p \wedge q_b^{j}.(x, y) = (-1250, 4000) \\
	& R_idO_jaW_5 \equiv dO^{(i,j)} \mathbf{U} (\neg q_b^{j}.p \wedge q_b^{j}.(x, y) = (-250, 4000) \\
	& R_idO_jaW_6 \equiv dO^{(i,j)} \mathbf{U} (\neg q_b^{j}.p \wedge q_b^{j}.(x, y) = (750, 4000) \\
	& R_idO_jaW_7 \equiv dO^{(i,j)} \mathbf{U} (\neg q_b^{j}.p \wedge q_b^{j}.(x, y) = (1750, 4000) \\
	& R_idO_jaW_8 \equiv dO^{(i,j)} \mathbf{U} (\neg q_b^{j}.p \wedge q_b^{j}.(x, y) = (2750, 4000) \\
	& R_idO_jaW_9 \equiv dO^{(i,j)} \mathbf{U} (\neg q_b^{j}.p \wedge q_b^{j}.(x, y) = (3750, 4000) \\
	& R_idO_jaW_{10} \equiv dO^{(i,j)} \mathbf{U} (\neg q_b^{j}.p \wedge q_b^{j}.(x, y) = (4750, 4000) \\
	\end{flalign*}
	
	However, after checking the satisfiability of those local missions, it is found out that the robot $R_{1}$ requires to move the objects $2$ (i.e. $?O_2Away$)  and $3$ (i.e. $?O_3Away$), and, similarly, the robot $R_4$ for its respective objects. Likewise, robots $R_2$ requires to move the object $1$, what robots $R_5$, $R_7$ and $R_9$ also requires for their corresponding objects. Thus, new local missions are generated in coordination layer considering those new assumptions such as $\tilde K^i_{MI}$ for robots $R_1$ and $R_4$ (i.e. $i = \{1, 4\}$) are:
	\begin{figure}[H]
		\centering
		\begin{tikzpicture}[shorten >=1pt,node distance=2.5cm,on grid,auto, bend angle=20, thick,scale=0.55, every node/.style={transform shape}]
		\node[state,initial] (s_0)   {};
		\node[state] (s_1) [right=of s_0] {};
		\node[state] (s_2) [right=of s_1] {};
		\node[state] (s_3) [right=of s_2] {};
		\node[state] (s_4) [right=of s_3] {};
		\node[state] (s_5) [right=of s_4] {};
		\path[->]
		(s_0) edge node [pos=0.5, sloped, above]{$?O_{i+1}Away$} (s_1)
		(s_1) edge node [pos=0.5, sloped, above]{$?O_{i+2}Away$} (s_2)
		(s_2) edge node [pos=0.5, sloped, above]{$R_ipO_i$} (s_3)
		(s_3) edge node [pos=0.5, sloped, above]{$R_idO_iaW_i$} (s_4)
		(s_4) edge node [pos=0.5, sloped, above]{$r_i$} (s_5);
		\end{tikzpicture}
		\vspace{-5 mm}
	\end{figure}    
	Equivalently, for robots $R_7$ and $R_9$ (i.e. $i = \{7, 9\}$):
	\begin{figure}[H]
		\centering
		\begin{tikzpicture}[shorten >=1pt,node distance=2.5cm,on grid,auto, bend angle=20, thick,scale=0.55, every node/.style={transform shape}]
		\node[state,initial] (s_0)   {};
		\node[state] (s_1) [right=of s_0] {};
		\node[state] (s_2) [right=of s_1] {};
		\node[state] (s_3) [right=of s_2] {};
		\node[state] (s_4) [right=of s_3] {};
		\path[->]
		(s_0) edge node [pos=0.5, sloped, above]{$?O_{i+1}Away$} (s_1)
		(s_1) edge node [pos=0.5, sloped, above]{$R_ipO_i$} (s_2)
		(s_2) edge node [pos=0.5, sloped, above]{$R_idO_iaW_i$} (s_3)
		(s_3) edge node [pos=0.5, sloped, above]{$r_i$} (s_4);
		\end{tikzpicture}
		\vspace{-5 mm}
	\end{figure}    
	Furthermore, besides to require to move other objects, the robots $R_2$ and $R_5$ respond a request to robots $R_1$ and $R_4$, respectively. Hence, these robots local mission (i.e. $i = \{2, 5\}$) are:
	\begin{figure}[H]
		\centering
		\begin{tikzpicture}[shorten >=1pt,node distance=2.5cm,on grid,auto, bend angle=20, thick,scale=0.55, every node/.style={transform shape}]
		\node[state,initial] (s_0)   {};
		\node[state] (s_1) [right=of s_0] {};
		\node[state] (s_2) [right=of s_1] {};
		\node[state] (s_3) [right=of s_2] {};
		\node[state] (s_4) [right=of s_3] {};
		\node[state] (s_5) [right=of s_4] {};
		\path[->]
		(s_0) edge node [pos=0.5, sloped, above]{$?O_{i+1}Away$} (s_1)
		(s_1) edge node [pos=0.5, sloped, above]{$R_ipO_i$} (s_2)
		(s_2) edge node [pos=0.5, sloped, above]{$!O_{i-1}Away$} (s_3)
		(s_3) edge node [pos=0.5, sloped, above]{$R_idO_iaW_i$} (s_4)
		(s_4) edge node [pos=0.5, sloped, above]{$r_i$} (s_5);
		\end{tikzpicture}
		\vspace{-5 mm}
	\end{figure}    
	Therefore, the robots $R_3$ and $R_6$ must respond to robots $R_1$ and $R_2$, and to robots $R_1$ and $R_2$, accordantly. Consequently, the local missions for these robots (i.e. $i = \{3, 6\}$)  are:
	\begin{figure}[H]
		\centering
		\begin{tikzpicture}[shorten >=1pt,node distance=2.5cm,on grid,auto, bend angle=20, thick,scale=0.55, every node/.style={transform shape}]
		\node[state,initial] (s_0)   {};
		\node[state] (s_1) [right=of s_0] {};
		\node[state] (s_2) [right=of s_1] {};
		\node[state] (s_3) [right=of s_2] {};
		\node[state] (s_4) [right=of s_3] {};
		\node[state] (s_5) [right=of s_4] {};
		\path[->]
		(s_0) edge node [pos=0.5, sloped, above]{$R_ipO_i$} (s_1)
		(s_1) edge node [pos=0.5, sloped, above]{$!O_{i-1}Away$} (s_2)
		(s_2) edge node [pos=0.5, sloped, above]{$!O_{i-2}Away$} (s_3)
		(s_3) edge node [pos=0.5, sloped, above]{$R_idO_iaW_i$} (s_4)
		(s_4) edge node [pos=0.5, sloped, above]{$r_i$} (s_5);
		\end{tikzpicture}
		\vspace{-5 mm}
	\end{figure}    
	Finally, the robots $R_8$ and $R_10$ must respond to robots $R_7$ and $R_9$, respectively. Hence, their new local missions (i.e. $i = \{8, 10\}$) are:
	\begin{figure}[H]
		\centering
		\begin{tikzpicture}[shorten >=1pt,node distance=2.5cm,on grid,auto, bend angle=20, thick,scale=0.55, every node/.style={transform shape}]
		\node[state,initial] (s_0)   {};
		\node[state] (s_1) [right=of s_0] {};
		\node[state] (s_2) [right=of s_1] {};
		\node[state] (s_3) [right=of s_2] {};
		\node[state] (s_4) [right=of s_3] {};
		\path[->]
		(s_0) edge node [pos=0.5, sloped, above]{$R_ipO_i$} (s_1)
		(s_1) edge node [pos=0.5, sloped, above]{$!O_{i-1}Away$} (s_2)
		(s_2) edge node [pos=0.5, sloped, above]{$R_idO_iaW_i$} (s_3)
		(s_3) edge node [pos=0.5, sloped, above]{$r_i$} (s_4);
		\end{tikzpicture}
		\vspace{-5 mm}
	\end{figure}    
	After synthesizing those local missions in the top-down layer and, consequently, the integrated task and motion plans in the global layer of the bottom-up layer, the robots start executing this plan. The Fig.~\ref{fig:map1001} shows the initial instant, when the robots $R_3$, $R_6$, $R_8$ and $R_10$ departed from the home position to the designed objects to pick up them. Other robots are requesting them to move the corresponding objects. Subsequently, these robots have responded the moving object away for corresponding requesting robot after these objects being picked up. This communication event is shown in the Fig.~\ref{fig:map1002} in the moment that robot $R_6$ responded to robot $R_5$ and $R_4$ and $R_5$ started to go pick its object up. When a robot crossed another robot way, the robot changes its trajectory reacting to the movement of the other robot every cycle time, as shown in the Fig.~\ref{fig:map1003}, \ref{fig:map1004} and \ref{fig:map1005}. Finally, the Fig.~\ref{fig:map1008} shows that all objects are dropped off to their designed position.
	
	\begin{figure}
		\centering
		\includegraphics[width=0.7\linewidth]{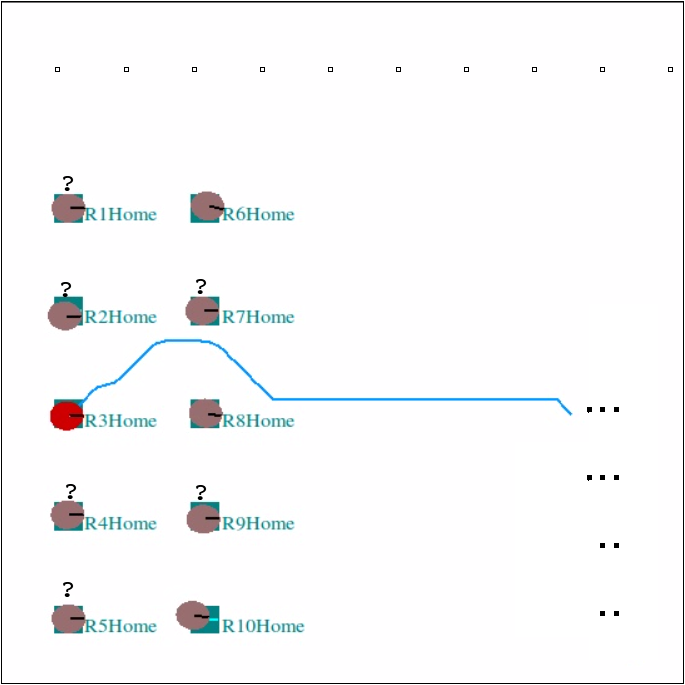}
		\caption{Initial instant of the scenario with 10 robots and 10 objects. The robots start in their home positions, and the red robot has his planned trajectory at this instant shown in blue line. The robots that are requesting an $OAway$ event have question marks above them. The filled dot lines in the bottom right are the objects. The others dots on top are the desired position of the objects specified in the global mission.}
		\label{fig:map1001}
		\vspace{-5 mm}
	\end{figure}
	
	\begin{figure}
		\centering
		\includegraphics[width=0.7\linewidth]{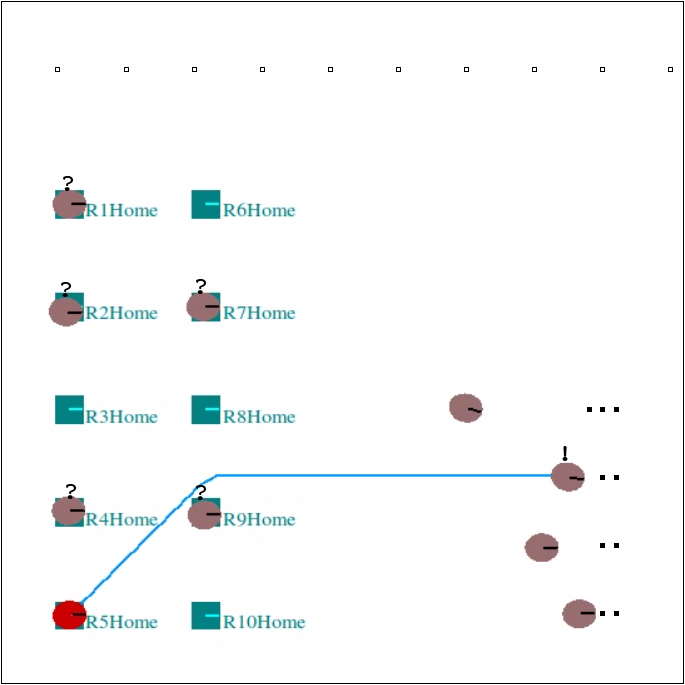}
		\caption{The instant that robot $R_6$ picked an object up and responded the $OAway$ event of robot $R_5$. The response is shown with a exclamation mark above the robot.}
		\label{fig:map1002}
		\vspace{-5 mm}
	\end{figure}
	
	\begin{figure}
		\centering
		\includegraphics[width=0.7\linewidth]{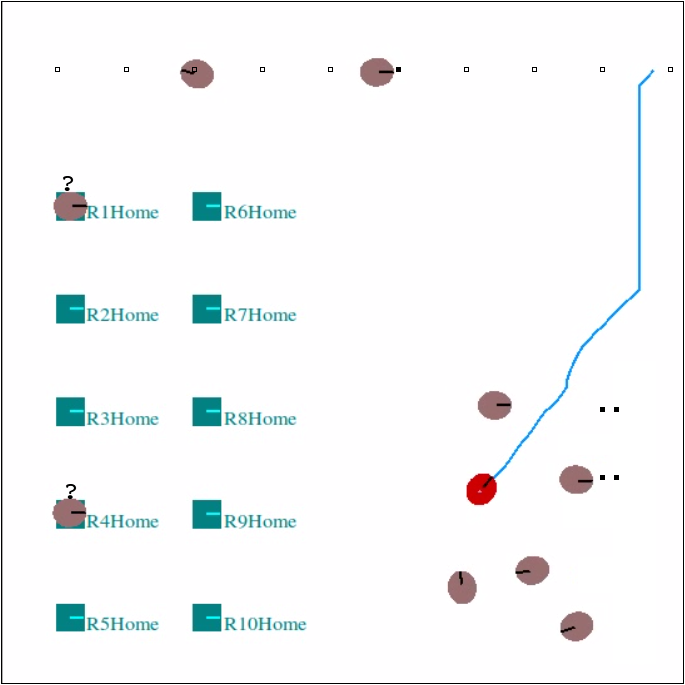}
		\caption{The robot $R_{10}$ in red is going to drop off position when the robot $R_2$ is in his way. }
		\label{fig:map1003}
		\vspace{-5 mm}
	\end{figure}
	
	\begin{figure}
		\centering
		\includegraphics[width=0.7\linewidth]{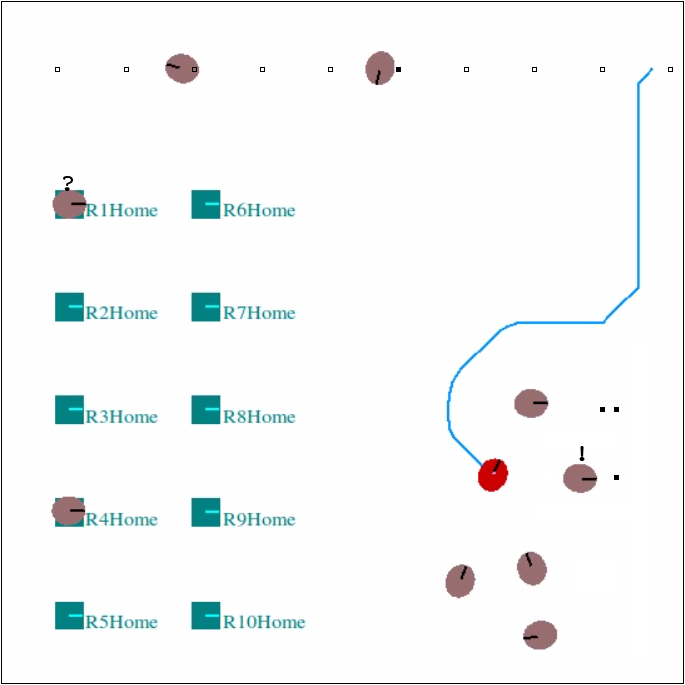}
		\caption{This instant shows the changes in the robot $R_{10}$ planned trajectory because of robot $R_{2}$ movements comparing with the instant at Fig.~\ref{fig:map1003}. The robot $R_{6}$ just drop its object off and is heading its home position, while the robot $R_{3}$ is arriving at its drop off position. Thee robot $R_{105}$ responded the $OAway$ event to robot $R_{4}$.}
		\label{fig:map1004}
		\vspace{-5 mm}
	\end{figure}
	
	\begin{figure}
		\centering
		\includegraphics[width=0.7\linewidth]{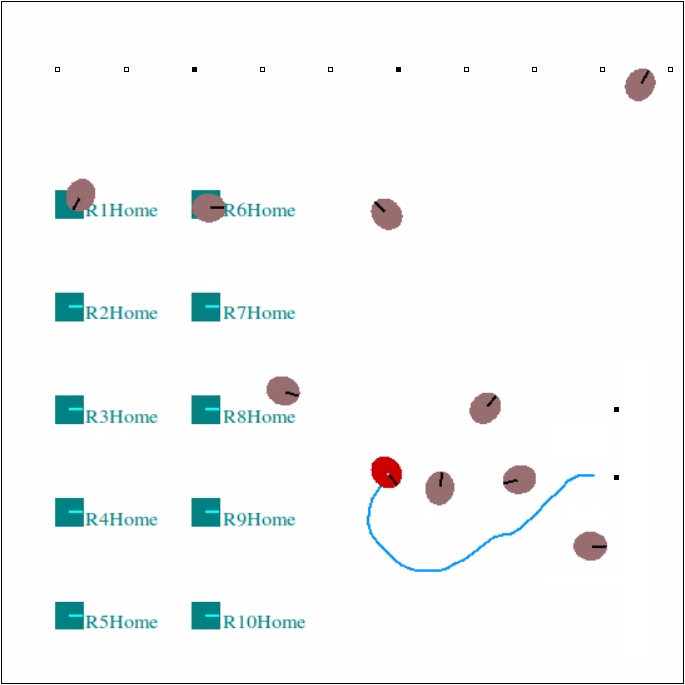}
		\caption{The robot $R_{4}$ is avoiding others robots in its way, while the robot $R_{10}$ is arriving to its drop off position. The robot $R_{3}$ is heading to its home position, where the robot $R_{6}$ have already arrived.}
		\label{fig:map1005}
		\vspace{-5 mm}
	\end{figure}
	
	\begin{figure}
		\centering
		\includegraphics[width=0.7\linewidth]{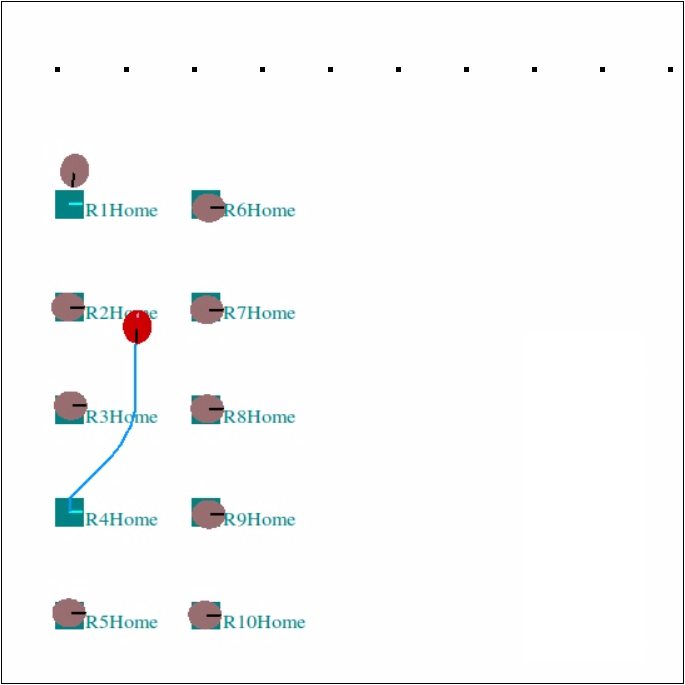}
		\caption{All objects where successfully left in their specified position, and the robots $R_{1}$ and $R_{4}$ are approaching to their home position.}
		\label{fig:map1008}
		\vspace{-5 mm}
	\end{figure}
\end{example}

\section{Conclusion}\label{sec:conclusion}
In this paper, we proposed a new framework in multi-agent system design by combining the formal top-down task decomposition and bottom up integrated task and motion planning (ITMP) approach CoSMoP in an iterative way.

Our unified framework can decompose the global mission into local missions based on which we synthesize the motion plan with pre-designed motion controllers that are proven to be safe (no active collision). Coordinations are added as necessary based on the feedbacks of CosMoP to guarantee the accomplishment of the global mission. The efficacy of the proposed method is shown in solving a warehouse example.



\bibliographystyle{IEEEtran}
\bibliography{cosmop,root}             

\end{document}